\documentclass[10pt,journal,compsoc]{IEEEtran}
\usepackage{stmaryrd}
\usepackage{adjustbox}
\usepackage{array}
\usepackage{multirow}
\usepackage{amsmath,bm}
\usepackage{amssymb}
\usepackage{pifont}
\usepackage{xspace}
\usepackage{tikz}
\usepackage[switch]{lineno} 
\usepackage{lipsum}
\newcommand*{\pd}[3][]{\ensuremath{\frac{\partial^{#1} #2}{\partial #3}}}
\newcommand{\HMC}[1]{{\textcolor{black}{#1}}}
\newcommand{\HMCrev}[1]{{\textcolor{black}{#1}}}

\newcommand{\cmark}{\ding{51}}%
\newcommand{\xmark}{\ding{55}}%

\newcommand{\best}[1]{\textcolor{red}{\textbf{#1}}}%
\newcommand{\sbest}[1]{\textcolor{blue}{\textbf{#1}}}%

\makeatletter
\usepackage{sidecap}
\def\@onedot{\ifx\@let@token.\else.\null\fi\xspace}
\DeclareRobustCommand\onedot{\futurelet\@let@token\@onedot}
\newcommand{\eg}{\emph{e.g}\onedot} 

\newcommand{\ie}{\emph{i.e}\onedot}

\newcommand{\etal}{\emph{et al}\onedot}
\makeatother
\ifCLASSOPTIONcompsoc
  \usepackage[nocompress]{cite}
\else

  \usepackage{cite}
\fi

\begin{document}
\title{Towards Partial Supervision for Generic Object Counting in Natural Scenes}
\author{Hisham Cholakkal$^{*}$, Guolei Sun$^{*\dag}$, Salman Khan, Fahad Shahbaz Khan,  Ling Shao and Luc Van Gool
\IEEEcompsocitemizethanks{\IEEEcompsocthanksitem * Hisham Cholakkal and Guolei Sun contribute equally to this paper.\protect\\
\IEEEcompsocthanksitem Hisham Cholakkal, Salman Khan and Fahad Shahbaz Khan  are with Mohamed Bin Zayed University of Artificial Intelligence, UAE.\protect\\
 \IEEEcompsocthanksitem Guolei Sun and Luc Van Gool are with the Computer Vision Laboratory,
ETH Zurich, Switzerland.\protect\\
\IEEEcompsocthanksitem Ling Shao is with Mohamed Bin Zayed University of Artificial Intelligence, UAE and Inception Institute of Artificial Intelligence, UAE.\protect\\
\IEEEcompsocthanksitem $\dag$ Guolei Sun is the corresponding author: guolei.sun@vision.ee.ethz.ch}}
\markboth{IEEE Transactions on Pattern Analysis and Machine Intelligence}%
{Cholakkal \MakeLowercase{\textit{et al.}}: Towards Partial Supervision for Generic Object Counting in Natural Scene}
\IEEEtitleabstractindextext{%
\begin{abstract}
Generic object counting in  natural scenes is a challenging computer vision problem. Existing approaches either rely on instance-level supervision or absolute count information to train a generic object counter. We introduce a partially supervised setting that significantly reduces the supervision level required for generic object counting. We propose two novel frameworks, named lower-count (LC) and reduced lower-count (RLC), to enable  object counting under this setting. Our frameworks are built on a novel dual-branch architecture that has an  image classification and a  density branch. Our LC framework reduces the annotation cost due to multiple instances in an image by using only lower-count supervision for all object categories. Our RLC framework further reduces the annotation cost arising from  large numbers of object categories in a dataset by only using lower-count supervision for a subset of categories and class-labels for the  remaining ones. The RLC framework extends our dual-branch LC framework with a novel weight modulation layer and a category-independent density map prediction.   
Experiments are performed on COCO, Visual Genome and PASCAL 2007 datasets. Our frameworks perform on par with state-of-the-art approaches using higher levels of supervision.  Additionally, we demonstrate the applicability of our LC supervised  density map  for image-level supervised instance segmentation. 
 \end{abstract}
}

\maketitle
\IEEEpeerreviewmaketitle
\IEEEraisesectionheading{\section{Introduction}\label{sec:introduction}}

Common object counting, also referred to as \textit{generic object counting}, seeks to enumerate instances of different object categories  in natural scenes. The problem is challenging as common object categories in natural scenes can vary from fruits to animals and counting must be performed  in both indoor and outdoor  scenes (\eg~COCO or Visual Genome datasets \cite{coco_eccv2014, krishnavisualgenome}). It is a highly valuable task for  scene understanding as it allows AI agents to summarize an image by quantifying instances from  each object category.  Moreover, object count can be used as an additional cue to improve other tasks, such as instance segmentation  and object detection.
 Existing works generally employ a localization-based strategy \cite{Girshick15ICCV,WhereAreBlobsECCV18} which locate the object first and then enumerate the located object instances, or utilize regression-based models \cite{Chattopadhyay_2017_CVPR}  for  directly predicting the object count.
 \begin{figure*}[t]
			\includegraphics[width=1\linewidth, clip=true, trim=0cm 0.1cm 2.5cm 0cm]{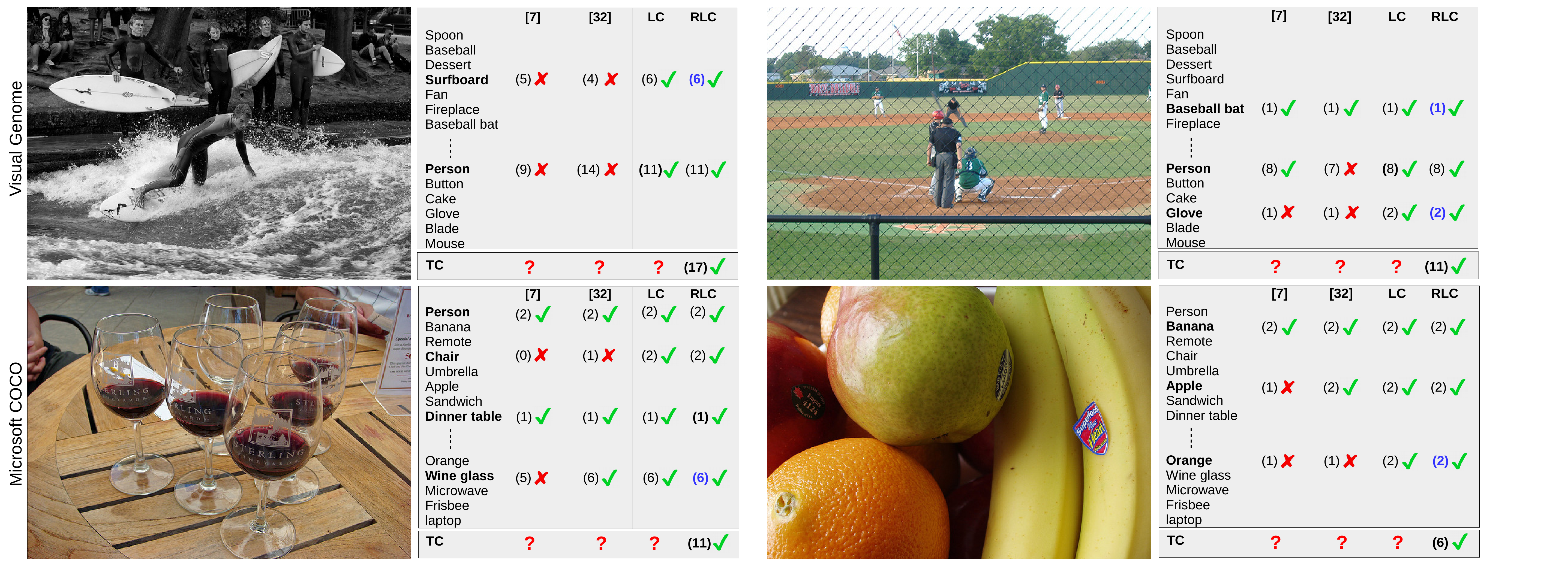}\\  \vspace{-0.5cm}
			\caption{Generic object counting with  partial supervision. Example results are shown on the Visual Genome (\emph{top}) and the COCO (\emph{bottom}) datasets. Due to large number of object categories in the dataset (609 for Visual Genome and 80 for COCO), acquiring accurate count annotations for natural scenes is laborious and costly.   
			We propose two settings to address this, where the first (LC)  reduces the annotation cost due to multiple instances and the second (RLC)  further reduces the annotation cost due to large numbers of object categories. While both frameworks reduce  supervision by only requiring image-level lower-count annotations, the LC framework requires these annotations
             for all categories, whereas RLC only needs them for a subset (\textcolor{blue}{blue} means the corresponding categories are not count-annotated during training of RLC framework).  Our two approaches (LC and RLC) significantly reduce the  annotation cost in comparison to the state-of-the-art instance-level {(LCFCN \cite{WhereAreBlobsECCV18}) and image-level (Glancing \cite{Chattopadhyay_2017_CVPR})} supervised methods. Here, the example counting results from   two datasets show the  generalizability of our frameworks to both  object counts beyond the  lower-count range ($>4$) and to  count-unannotated categories. Due to  the absence of category-specific counts for some classes, we introduce an  additional  category-agnostic total count  measure in our RLC framework to facilitate generalization   across categories and   provide accurate category-agnostic total count (TC) predictions. (correct/incorrect/unavailable predictions are marked with \textcolor{green}{\cmark}/ \textcolor{red}{\xmark}/\textcolor{red}{\textbf{?}} respectively). 
						}
			\label{fig:CountingIntro}
\end{figure*}
	
 Most localization  or  regression-based generic object counting approaches \cite{WhereAreBlobsECCV18, Chattopadhyay_2017_CVPR} require manual annotation of each instance, either in the form of a bounding box or  a point (\eg the center of the object). These instance-level annotations are time consuming since they need to be sequentially marked at each instance. The cost of user-intensive, instance-level annotation can be reduced by utilizing image-level supervision, which only  requires  the instance count of each object category in an image \cite{Chattopadhyay_2017_CVPR}. However, for large object counts,  even image-level annotation is laborious since it requires the annotator to sequentially focus on each object instance. Further, the annotation time increases with the number of object categories to be enumerated. Therefore, existing generic object counting approaches, utilizing either instance or image-level annotations are not scalable for natural scenes that have multiple instances of a large number of object categories (\eg~ the Visual Genome dataset \cite{krishnavisualgenome}).

  In this work, we propose a partially supervised setting (see Fig.~\ref{fig:CountingIntro}) that reduces the annotation effort further than image-level count supervision. Under this partially supervised setting, we reduce the annotation cost due to (i) multiple instances and (ii) large numbers of categories. To reduce the annotation cost due to multiple instances, we propose a novel LC framework based on a \textit{lower-count (LC)} supervision strategy. The LC supervision strategy is motivated by the psychological studies \cite{psychological_study1,psychological_study2,psychological_study3,psychological_study4}  suggesting that humans are capable of counting objects non-sequentially using holistic cues for fewer object counts, termed as the subitizing range or lower-count (generally 1-4). We utilize this property in our LC framework by only using object count supervision within the lower-count range. The proposed LC framework consists of an image classification and a density branch. The image classification branch estimates the presence or absence of objects, whereas the density branch predicts the category-specific  object  count. In addition, the density branch also predicts the  spatial  distribution of object instances for each object category, by constructing separate  density  map  for  each  category. This  spatial distribution is later shown to be useful for delineating spatially adjacent object instances in weakly supervised instance segmentation.

The LC supervised setting, introduced above, considers lower-count annotations ($\leq$4) for ``all object categories".  Since there is often a  large number of  object categories appearing in  natural scene datasets (\eg~ several hundred  categories in the Visual Genome dataset ~\cite{krishnavisualgenome}), even obtaining LC annotations for all of these object classes is cumbersome. This necessitates the development of an object counting approach that can further reduce the supervision  level and  generalize accurately to novel object categories.  With this objective, we propose a more challenging partially supervised setting for object counting, where image-level classification labels are available for all categories, but lower-count annotations are known for ``only a subset" of object categories. We call this setting \textit{reduced lower-count (RLC)} supervision.  Since the counting operation looks for {repetitive patterns in an image}, we hypothesize that an object   counter will generalize to categories for which no count annotations are given, enabling object counting  under this RLC setting. To address this challenging setting, we  propose an RLC framework that extends our dual-branch (image classification and density) LC framework with the following changes.       
First, we introduce a novel weight modulation layer that effectively transfers knowledge from count-annotated categories to those categories without any count annotation. To further support the generalization of count predictions to these  count-unannotated categories, an additional sub-branch is introduced to the density branch for predicting the total count of objects in natural scenes, irrespective of their category. 

The LC framework leads to  better-quality density maps that can be used for other applications, such as instance-segmentation, whereas the RLC framework scales up  object counting to novel categories whose count annotations are not available during training.  
Fig.~\ref{fig:CountingIntro} shows that our partially supervised approaches (LC and RLC) are able to accurately predict object counts beyond the lower-count range.  For example, although the count annotations are not available for categories like  'surfboard', and 'wine glass', their instance counts beyond the lower-count range are predicted accurately by our RLC framework (see images on the left in Fig.~\ref{fig:CountingIntro}). Despite the presence of multiple categories without any count annotations,  our RLC framework accurately predicts  the total count (TC) of objects in these images.

\textbf{Contributions:} In this work, we introduce a partially  supervised setting  that can substantially reduce the required level of supervision for generic object counting in natural scenes. Our contributions are as follows: \textbf{(a)} We propose two novel frameworks (LC and RLC)  for generic object counting. Our LC framework aims at reducing the annotation cost due to multiple instances while the RLC framework targets reducing the annotation cost due to large numbers of object categories and instances in natural scene datasets. The LC framework comprises an image classification and density branch. The RLC framework extends our dual-branch LC framework with a novel weight modulation layer that contributes to the generation of category-specific density maps and is generalizable to categories without count annotations.  Moreover, a category-agnostic sub-branch is introduced in the density branch of the proposed RLC framework to  generate a category-independent density map and therefore estimates  accurate \textit{total object count} for an image.
\textbf{(b)}
 We propose LC supervised training of density maps for generic object counting, and demonstrate its applicability for image-level supervised instance segmentation. \textbf{(c)} To the best of our knowledge, we are the first to introduce a generic object counting approach that targets the transferability of count prediction to the RLC supervised scenario (where  only image-level category labels are available,  without count-annotations). 
 \textbf{(d)} We extensively evaluate our approach on multiple datasets, such as Visual Genome~\cite{krishnavisualgenome} and MS-COCO~\cite{coco_eccv2014}. Our approach obtains state-of-the-art performance for category-specific object counting on these datasets. Moreover, our RLC framework  demonstrates impressive results for total and category-specific counts,  even in the presence of several categories without any count annotations.

 \section{Related work}
 
  \indent\textbf{Generic object counting in natural scenes:}  Object counting methods in the literature can be categorized as the ones that provide class-wise object counts (\emph{category-specific} counters) and those that provide a total count for all objects (\emph{category-independent} counters). 
Category-specific object counting in natural scenes has been recently investigated \cite{Chattopadhyay_2017_CVPR,WhereAreBlobsECCV18}.  Chattopadhyay \etal\cite{Chattopadhyay_2017_CVPR} proposed  object counting strategies that use instance-level and image-level  supervisions. 
 The instance-level (bounding box)  supervised strategy, denoted as subitizing, estimates a large number of objects by dividing an image into non-overlapping regions, assuming the object count in each region falls within the subitizing, or lower-count, range.  
The image-level (per-category count) supervised strategy, denoted as  glancing, uses a  regression loss to train a convolutional neural network that can  predict per-category object counts.  \HMC{Similarly, \cite{GenericDivideCount} proposes to divide an image into different regions for obtaining the object count.  It employs inclusion-exclusion principle from set theory and enforce consistency in counts when dealing with overlapping image regions. Both these approaches \cite{Chattopadhyay_2017_CVPR, GenericDivideCount} use count annotations from both within and beyond the subitizing or lower-count range to predict the per-category count of each object, without providing information about their location.} 
In contrast, our approach requires neither  box annotations nor count information beyond the lower-count range. The work of \cite{WhereAreBlobsECCV18} proposed a  point-level supervised approach that requires post-processing steps such as \cite{connectedComponent} to estimate object counts.  Our method does not require such post-processing steps and  instead directly predicts the  object counts by simple summation of  the  density maps. Aside from category-specific counting, class-agnostic or category-independent object counting  was previously investigated for salient object subitizing in \cite{sos_subitizing_cvpr2015,he2017delving, Islam_2018_CVPR}, where the category-agnostic counting of salient objects was performed only  within the subitizing or lower-count range.  In contrast, our approach     estimates  the object counts both within and beyond the lower-count range.  

Another strategy for common-object counting is to count the number of instances detected by an object detector\cite{Girshick15ICCV}. Although these object detectors are trained for precise localization of object boundaries with bounding-box supervision, the number of instances detected may not match with the ground-truth count, especially at large ground-truth counts \cite{Chattopadhyay_2017_CVPR}.  
This is  mainly due to the fact that optimizing the network for precise localization of each object boundary  may result in false negative detections  on challenging images with heavy occlusion,  and hence under-counting \cite{Chattopadhyay_2017_CVPR}. Adjusting the detector parameters (such as the non-maximum suppression threshold) to address this issue may  lead to false positives and hence  over-counting. Our method has dedicated count loss terms which enable accurate count  predictions under these challenging scenarios. 

 \indent\textbf{Crowd counting in surveillance scenes:}
 \HMC{Crowd counting approaches in surveillance scenes  \cite{cheng2019learning, Cheng_2019_ICCV, OpenSetCloseSet_ICCV2019, CAN_2019_CVPR, Shi_2019_CVPR, SyntheticData_2019_CVPR, Liu_2019_CVPR, CSRnetDialatedConv_2018_CVPR,Sindagi_2019_ICCV, crowdCountingDetection1,crowdCountingDetection2,Wan_2019_CVPR,Liu_2019_CVPR_attention} generally aim at counting large numbers of instances from one or few object categories.}  Most crowd counting approaches \cite{CSRnetDialatedConv_2018_CVPR, cheng2019learning,CAN_2019_CVPR,rankingUnlabelleddata_CVPR2018, SyntheticData_2019_CVPR, Cao_2018_ECCV },  require instance-level supervision and use a density map to  preserve the spatial distribution of people in an image,   obtaining  the person count for the image by summing over the predicted density map.  Additional unlabelled data and synthetic data are  used in  \cite{rankingUnlabelleddata_CVPR2018}, \cite{SyntheticData_2019_CVPR}, respectively to  improve crowd counting. \HMC{The work of \cite{von2016gaussian} propose a weakly supervised  strategy that requires only region-level count annotations, instead of instance-level point annotations, for pedestrian counting in surveillance scenes and cell counting for medical applications.}

\HMC{Recent crowd counting methods focus on developing novel attention mechanisms \cite{Liu_2018_CVPR, Liu_2019_CVPR_attention, Zhang_2019_ICCV, jiang2020attention, attention_AAAI},  generation of density maps \cite{Ma_2019_ICCV,Wan_2019_ICCV,Cao_2018_ECCV, CAN_2019_CVPR,Xu_2019_ICCV}, and  context or scale-aware designs \cite{Liu_2019_ICCV,Xu_2019_ICCV,Liu_2019_CVPR_context}.}  
\HMC{A detailed survey of recent CNN-based crowd counting approaches is available at \cite{gao2020cnn}}. 
A fully convolutional framework with multiple dilated convolution kernels was  proposed in \cite{CSRnetDialatedConv_2018_CVPR}. These dilated convolution kernels increase the receptive fields of  deep feature maps, without losing spatial resolution through pooling operations, enabling counting under highly congested scenes. \HMC{Goldman \etal~\cite{goldman2019precise} propose a detector based strategy for counting objects in densely packed surveillance scenes such as  retail stores or car parking areas.} 
 Lu \etal \cite{Lu18} proposed a generic matching  network that counts the instances of a given object category,  using an exemplar patch containing the object  of interest.  This method uses an image-matching strategy to obtain the  instance count of an object category represented by a training exemplar. However, the exemplar need to be similar to the object instances in a test image, hence less accurate  for  generic object counting in natural scenes having large  intra-class variations.

In this paper, we address the problem of large-scale common or generic object counting,  where the objective is to enumerate instances from a large number of object categories. Although the number of instances in an  individual category is  smaller compared to crowd counting, accurately predicting  the object count is highly challenging due to  the presence of a large number of diverse object categories in the same image. Moreover, dense occlusions, cluttered regions and 
large intra-class variations in natural scenes increase the complexity of the problem (see Fig.~\ref{fig:CountingIntro}).  We propose  partially supervised  (LC or RLC)   density map estimation approaches for generic object counting in natural scenes, which predict the category-specific object count while preserving the spatial distribution of objects. Fig.~\ref{fig:densitymapsLC_RLC} shows examples of the category-specific density maps of natural scenes generated by our partially supervised LC and RLC frameworks.  
 \begin{figure*}[t]
		\centering
			\includegraphics[width=0.95\textwidth, clip=true, trim=0cm 18cm 7.2cm 0cm]{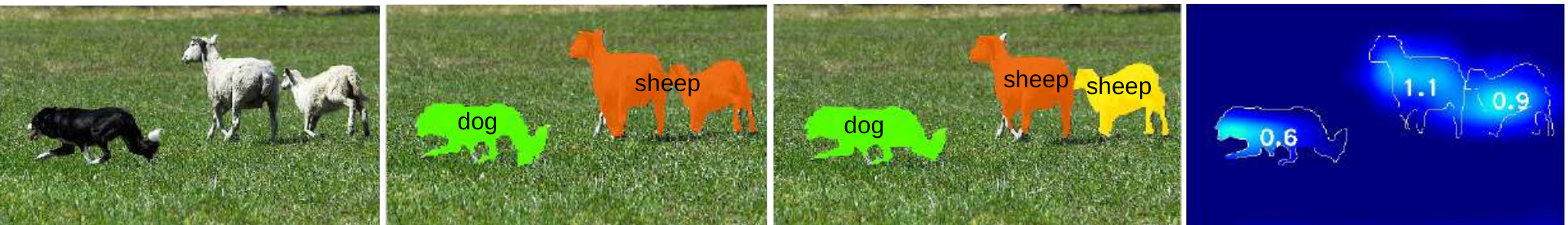}\\
				\centering
				\includegraphics[width=0.95\textwidth, clip=true, trim=0cm 14cm 7.2cm 3.2cm]{introduction_segmentation3.pdf}\\
				\vspace{-0.5cm}
			\hspace*{0.02\linewidth} (a) Input Image \hspace*{0.10\linewidth} (b) PRM \cite{PRM}\hspace*{0.11\linewidth}(c) Our Approach \hspace*{0.07\linewidth} (d) Our Density Map \\  
			\caption{Examples showing the category-specific density maps, generated by our LC framework, and their usability  for improving image-level supervised instance segmentation. We show instance segmentation results using the PRM method \cite{PRM} (b) and our approach (c), on  PASCAL VOC 2012 images. Top row: The PRM approach \cite{PRM} fails to delineate two spatially adjacent  sheep category instances. Bottom row: Single person parts are predicted as multiple persons along with inaccurate mask separation results in over-prediction (7 instead of 5) by the PRM method \cite{PRM}. The per-category  density maps (d) obtained in the lower-count supervised (LC) setting provides the  spatial distribution of object count;  hence	the accumulation of the  density map over a local spatial region generally  indicates the object count in that region.  This property is used to   penalize the instance mask predictions containing multiple object instances (object count $\ge2$)  or part of objects (object count $\approx0$), and hence to  improve  weakly supervised instance segmentation (c). The density map accumulation for each predicted mask is shown inside the contour, drawn for clarity. In the top row, the density maps for the sheep and dog categories are overlaid.}             
					\label{fig:SegmentIntro}
			\vspace*{-0.2cm}
\end{figure*}

\indent\textbf{Generic object counting with limited supervision:} 
In this work, our aim is to address the problem of generic object counting with limited supervision. Our RLC framework is motivated by the transfer function  used for instance-mask prediction in \cite{hu2018learning} and  object detection in \cite{hoffman2014lsda}. Different from these, however, our weight modulation layer modifies the convolution weights of an image classifier branch to estimate   category-specific object counts. As a result, the proposed layer can be learned from an entire image, without using instance-level annotations or pooling as in \cite{hu2018learning, hoffman2014lsda}. Although several weak supervision strategies, such as  zero-shot learning,  semi-supervised learning and weakly supervised learning \cite{Gao_2018_ECCV}, have been employed for applications such  as object detection \cite{rahman2018zeroshot} and semantic segmentation \cite{xu2015learning},  they have not been employed for generic object counting in natural scenes.   
Zero-shot learning  approaches \cite{rahman2018zeroshot}   generally require a natural language description of unseen categories,  while few-shot learning and semi-supervised learning approaches \cite{snell2017prototypical,Zhao_2018_ECCV} require a minimum amount of training data for every category.  Different to these approaches, our RLC framework  adopts the convolution weights  learned for the image classification task to obtain the object count of categories without count annotations.   \HMC{The generation of kernel weights is previously investigated  for various applications in hypernetworks \cite{HyperNetICLR2017}, dynamic convolution \cite{klein2015dynamic}, dynamic filtering \cite{de2016dynamic}, and adaptive convolution \cite{kang2017incorporating}. Instead of generating  the kernel weights using CNN features, our weight modulation layer is a simpler transfer function that adapts the  convolution weights between two tasks, \ie, the weights trained for classification task are adapted for  counting task.}\\
\indent\textbf{Differences from the preliminary version \cite{CommonCounting_CVPR19}:} This paper is an extension of our preliminary
work published at the  CVPR conference in 2019 \cite{CommonCounting_CVPR19}. The main contributions of this longer version compared to its preliminary work are the following.  The LC supervised setting introduced in the preliminary version \cite{CommonCounting_CVPR19} requires counts within the lower-count range for `\emph{all}' object categories. Here, we extend this to  an even more challenging setting where counts within the lower-count range are known for only a subset of categories during training (RLC supervision). For count prediction in the  RLC setting, we introduce a novel weight modulation layer that transfers knowledge from objects with annotated counts to the cases where count information is unknown. Further, the RLC framework introduced in this paper can also predict category-agnostic total object counts,  whereas the approach proposed in the preliminary version \cite{CommonCounting_CVPR19} can only predict category-specific object counts. Moreover, our partially supervised RLC problem setting enables object counting on large-scale datasets that have  hundreds of object categories (e.g., Visual Genome), where  count annotation for all object categories is difficult to obtain.  Compared to \cite{CommonCounting_CVPR19}, additional ablation experiments for the LC framework \cite{CommonCounting_CVPR19} are performed on the  COCO dataset  and  performance is evaluated on an additional large-scale dataset with a high number of object categories (Visual Genome). Finally, we also perform a thorough evaluation of the proposed RLC framework on the COCO and Visual Genome datasets. 
\section{Generic Object counting with partial  supervision}
\begin{figure}[t]
			\includegraphics[width=1\columnwidth, clip=true, trim=1cm 1.3cm 28.5cm 0cm]{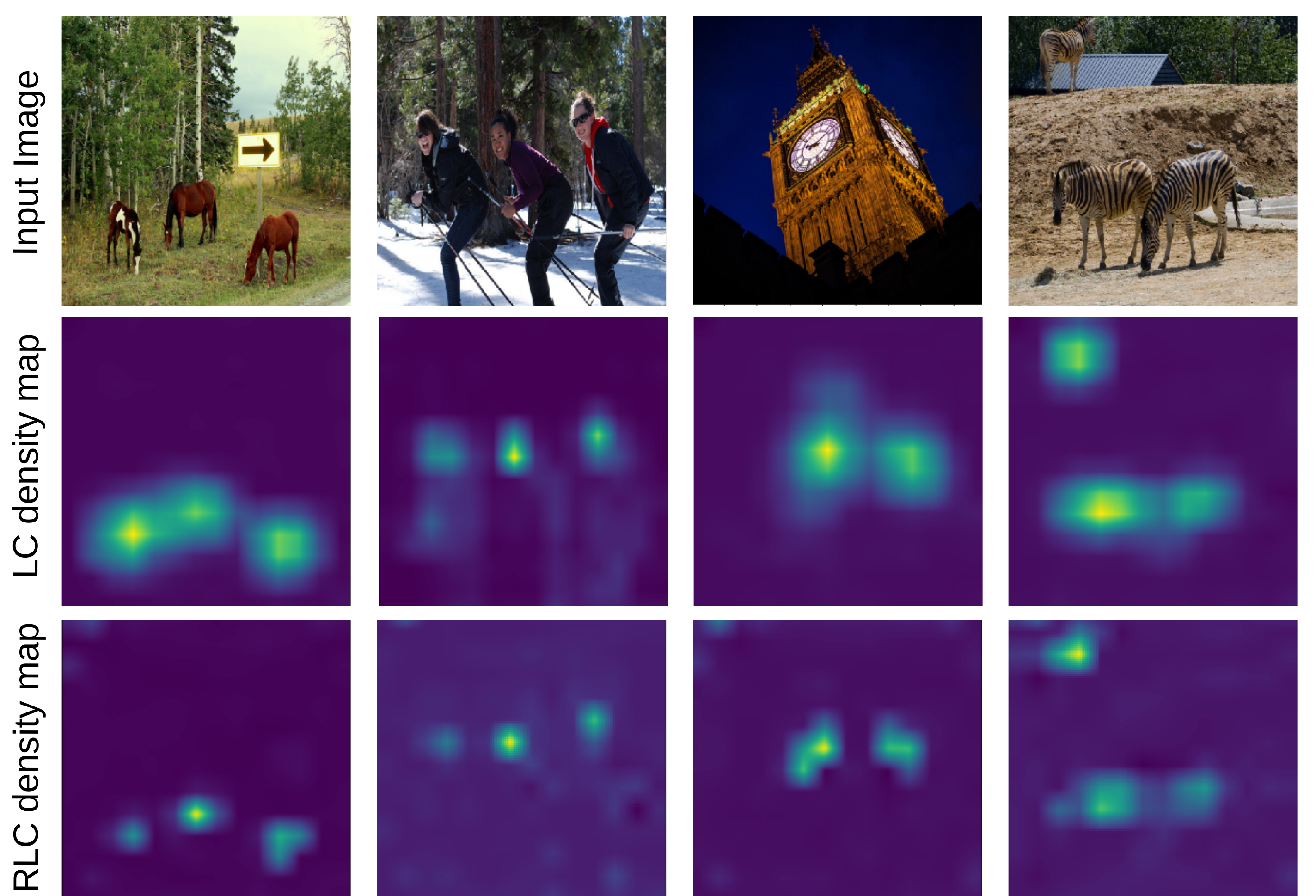}\\  \vspace{-0.50 cm}	
			\caption{Example category-specific density maps produced by our LC and RLC frameworks for horse, person, clock and zebra categories, respectively, on images from COCO dataset. Despite being trained using image-level lower-count supervision, the spatial distributions of objects are preserved in both the LC and RLC density maps. Note that,  in the case of our RLC framework,  only category-level annotations (no lower-count annotations) were available  for the clock and zebra categories.}
			\label{fig:densitymapsLC_RLC}
			 \vspace*{-0.0cm}
\end{figure}

As mentioned earlier,  obtaining object count annotations for natural scenes is expensive due to the presence of: (i) multiple instances, and (ii) a large number of object categories that need to be enumerated.  To address these two challenges,  we introduce two partially supervised settings with the  aim of significantly reducing the annotation cost beyond image-level count annotation.
 In the first setting, named  lower-count (LC) supervision, exact counts  are annotated \textit{only until four}, for all object categories, and hence the annotation cost is reduced on images that have a  large number of instances. In the second setting, named  reduced lower-count (RLC) supervision, we further reduce the supervision  such that the LC supervision is only provided   for a subset of object categories, and the remaining ones are simply  annotated with  category labels 
 (\ie, binary labels indicating the presence  or absence of object categories).
 This helps in reducing the annotation costs due to   multiple instances and large numbers of object categories.   
To facilitate object counting under these partially supervised settings (LC and RLC), we  introduce a novel dual-branch architecture,  where the first branch, named the 'image classification branch',  predicts the presence or absence of object categories,  while the second branch, named the 'density branch',  produces a density map for accurately predicting the object count. Fig.~\ref{fig:densitymapsLC_RLC} shows examples of the density maps produced by our LC and RLC frameworks. Specifically,  Fig.~\ref{fig:densitymapsLC_RLC} demonstrates that the category-specific density maps produced by our LC framework better preserve spatial information and can be used for object localization applications, such as  weakly supervised instance segmentation (see Fig.~\ref{fig:SegmentIntro}). 

 As discussed earlier, it is easier to obtain LC supervision when there is a lower number of object categories (\eg~20 categories in the PASCAL VOC dataset). 
 The annotation effort increases with the number of object categories  and it is thus difficult to obtain LC supervision when the number becomes too large (\eg~ several hundred categories in Visual Genome dataset).  
 In such cases, 
 our RLC framework is more suitable since it only requires LC supervision for a subset of categories and  category-level annotations for the remaining ones.

 	\begin{figure*}[t]
		\centering
						\includegraphics[width=1.0\linewidth, clip=true, trim=0cm 38cm 0.0cm 0cm]{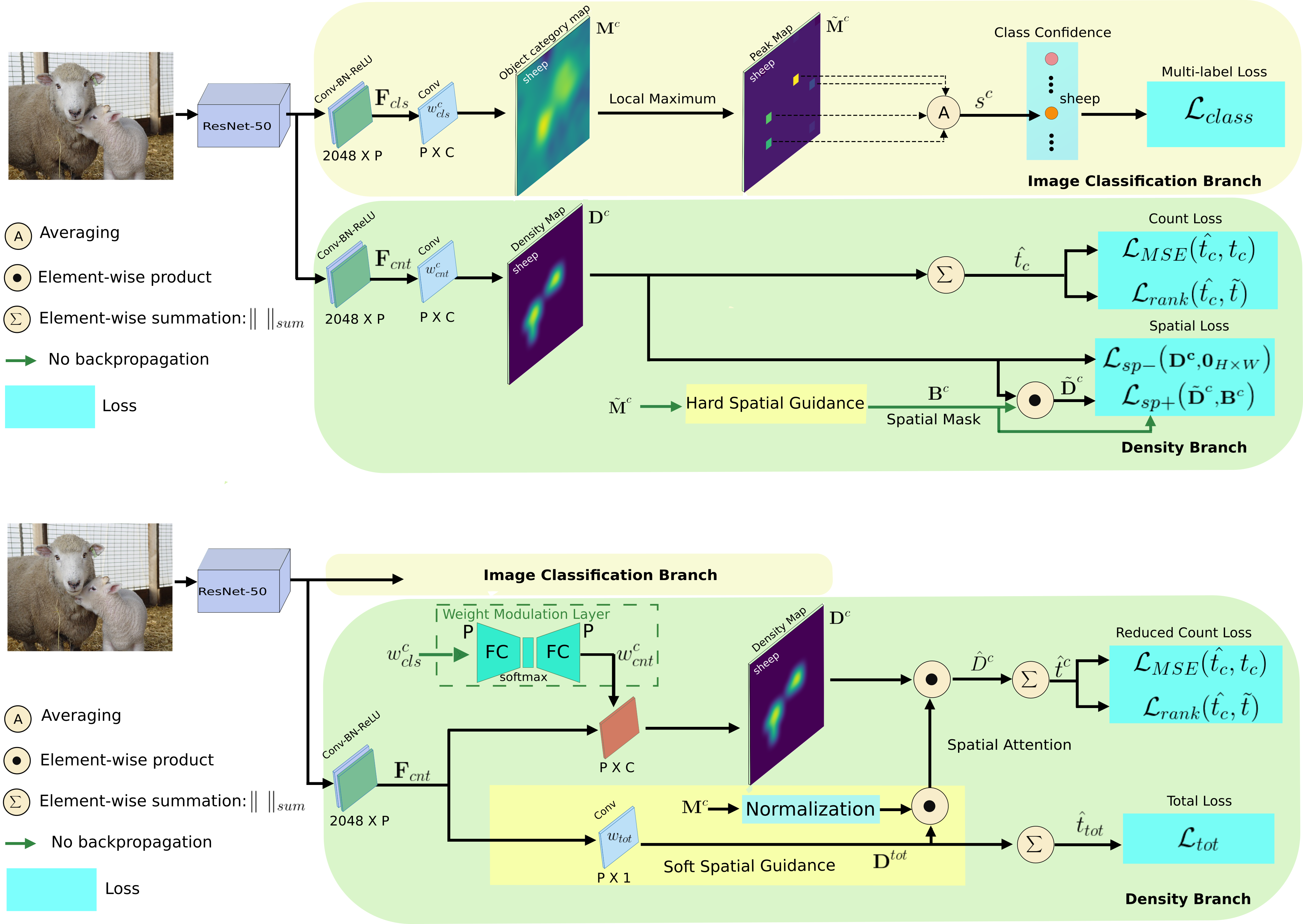}\\ \vspace{-0.15cm}
				\caption{Overview of our LC architecture with image classification and density branches which are trained jointly using lower-count (LC) supervision. The image classification branch predicts the presence or absence of objects. This branch is used to generate a spatial mask for training the category-specific  density branch. The density branch has two terms (spatial and count) in the loss function and produces a category-specific density map to predict the category-specific object count and preserves the spatial distribution of objects. }
			\label{Fig:architectue}
		\end{figure*}

Important  notations related to our partially supervised experimental settings are shown  in Table~\ref{table:notations}. 
In the next section, we introduce our novel framework for LC supervision,  followed by its extension to the   RLC framework in Sec.~\ref{SPC_overview}.

\begin{table}[t]
\resizebox{\linewidth}{!}{%
\begin{tabular}{|l l|}
\hline
$I$ & Training image \\
 $S_0$ & Set of categories which are absent  in an image $I$  \\
 $S$ & Set of categories in the image  $I$ that have  counts \\
  &  within the lower-count range (\ie, $1\leq t_c < \tilde{t}$) \\
$\tilde{S}$& Set of categories  in the image  $I$ that have  counts  \\
&   beyond the lower-count range  (\ie, $t_c \ge \tilde{t}$)   \\
C & Total number of object categories in a given dataset \\
$\mathcal{A}$ & Set of categories in the dataset that have  both  \\
& count and category annotations \\
$\mathcal{B}$ & Set of categories in the dataset that have only \\
&  category annotations \\
$\textbf{t}$ & A vector indicating the ground-truth count of \\ 
&  all $C$ object categories\\ 
$t_c$ & Ground-truth count of object category $c$\\ 
$\hat{t}_c$ & Predicted count of object category $c$\\ 
$\tilde{t}$ & Smallest count for $\tilde{S}$, \ie~5\\ 
$t_{tot}$ & Ground-truth category-agnostic  total count in an image $I$ \\ 
$\hat{t}_{tot}$ & Predicted category-agnostic total count in an image $I$ \\
 \hline
\end{tabular}%
}\vspace{0.3em}
\caption{Notations used  in our partially supervised (LC and RLC) frameworks.}
\label{table:notations}
\vspace{-0.5cm}
\end{table}

\section{Lower-count Supervised Framework}
\label{proposedmethod_start}

 Let $\textbf{I}$ be a training image and $\textbf{t}=\{t_1, t_2, ..., t_c, ..., t_C\}$ be the corresponding vector for the ground-truth count of $C$~object categories. Instead of using an absolute object count, we employ a lower-count strategy to reduce the amount of image-level supervision. Given an image $\textbf{I}$, object categories are divided into three  non-overlapping sets based on their respective instance counts. The first set, $S_0$, indicates object categories which are absent in $\textbf{I}$ (i.e., $t_c=0$). The second set, $S$, represents categories within the lower-count range (i.e, $0<t_c \le 4$). The final set, $\tilde{S}$, indicates categories beyond the lower-count range (i.e., $t_c\ge \tilde{t}$, where $\tilde{t}=5$). Next, we explain the proposed network architecture  (LC architecture) for the LC supervised setting.

\noindent\textbf{The Proposed LC Architecture:}
Our approach is built upon an ImageNet pre-trained network backbone (ResNet50) \cite{ResNet}. The proposed network architecture has two output branches: the image classification and  density branches (see Fig.~\ref{Fig:architectue}). The  image classification branch estimates the presence or absence of objects.    The  density branch predicts the category-specific object count and the spatial distribution of object instances for  each object category, by constructing a separate density map for each  category. We remove the global average pooling layer from the backbone and preserve the spatial information from the backbone features. The resulting backbone features with $2048$ channels are used as a common input to both the image classification and  density branches. 
We then add two  $1\times 1$ convolutions in each branch, where the  first convolution has  $P$ output channels and the second convolution  has $C$ output channels,  resulting  in  a  fully  convolutional  network~  \cite{FCN}.  Here, $C$ is the number of object categories and $P$ is empirically set to be proportional to $C$. These convolutions  are separated   by a batch normalization and a ReLU layer. 
The input features for the last convolution layers of the image classification branch and the density branch are indicated as $\mathbf{F_{cls}}$ and $\mathbf{F_{cnt}}$, respectively (see  Fig.~\ref{Fig:architectue}). The last convolution layers in the image classification and density branches output $C$  spatial maps (corresponding to $C$  object categories),  named as object  category maps and density maps, respectively. The object category maps help  in object localization, while the density maps are useful for estimating the object count  in a given spatial region. \\
\noindent\textbf{The Proposed Loss:}
\label{subsec:ProposedLossFn}
Let $\mathbf{M}=\{\mathbf{M}^c {\in}\mathbb{R}^{H\times W}: c{\in}[1,C]\}$ denotes the object category maps in the image classification branch and $\mathbf{D}=\{\mathbf{D}^c {\in}\mathbb{R}^{H\times W}: c{\in}[1,C]\}$ represents category-specific density maps produced by the  density branch. 
 Here,
 $H\times W$ is the spatial size of both the object category and category-specific density maps. The image classification and  density branches are jointly trained, in an end-to-end fashion, given only LC supervision with the following loss function:
\begin{equation} 
\label{eq:maineq}
     \mathcal{L}=  {{\cal{L}}_{class}}+\underbrace{{\cal{L}}_{spatial}+{\cal{L}}_{count}}_{\text{Category-specific~density~branch}}.
   \end{equation}
Here, the first term refers to  the multi-label image classification loss \cite{multilabelsoftmargin} (see Sec.~\ref{peakstimulation}).  The last two terms, ${\cal{L}}_{spatial}$ and ${\cal{L}}_{count}$, are used to train the  density branch (Sec.~\ref{density}). 

\subsection{Image Classification Branch}
\label{peakstimulation}
Generally, training a density map requires instance-level supervision, such as point-level annotations \cite{OxfordDensityNIPS2010}. Such information is unavailable in our LC supervised setting. To address this issue, we propose  generating a spatial mask   by exploiting the coarse-level localization capability of an image classifier \cite{oquab2015object, CAM} via object category maps.  These object category maps are generated from a fully convolutional architecture, shown in Fig.~\ref{Fig:architectue}. 

While specifying classification confidence at each image location, class activation maps (CAMs) struggle to delineate multiple instances from the same object category \cite{oquab2015object, CAM}. Recently, the local maxima of CAMs were further boosted to produce object category maps, during an image-classifier training for image-level supervised instance segmentation \cite{PRM}. The boosted local maxima  aim at falling on distinct object instances. For details on boosting local maxima, we refer to \cite{PRM}. Here, we use local maxima  locations to generate a spatial mask, which is used as a pseudo ground-truth  for training the density branch (Sec.~\ref{density}).

As described earlier, object categories in $\textbf{I}$ are divided into three non-overlapping sets: $S_0$, $S$ and $\tilde{S}$. To train a one-versus-rest image classifier, we derive
binary labels from $t_c$ that indicate the presence ($c\in\{S, \tilde{S}\}$) or absence  ($c \in S_0$) of object categories. Let the peak map $\tilde {\textbf{M}}^{c}  \in R ^{H\times W}$ be derived from the $c^{th}$ object category map ($\textbf{M}^{c}$) of $\textbf{M}$, by using local maxima locations such that:
 \vspace{-0.1cm}
\begin{equation*}
\small
 \tilde{\textbf{M}}^{c}(i,j)=\begin{cases}
    \textbf{M}^{c}(i,j),&\text{if~ $\textbf{M}^{c}(i,j)>\textbf{M}^{c}(i-r_i,j-r_j)$},\\
    0, & \text{otherwise}.
  \end{cases}
   \vspace{-0.0cm}
\end{equation*}
Here,  $-r \leq r_i \leq r$ and   $-r \leq r_j \leq  r$, where  $r$ is the radius for the local maxima computation. We set $r=1$, as in \cite{PRM}. The local maxima are searched at all spatial locations with a unit stride.  
To train an image classifier, a class confidence score $s^c$ of the $c^{th}$ object category is computed as the average of non-zero elements of $\tilde{\textbf{M}}^{c}$. In this work, we use the multi-label soft-margin loss \cite{multilabelsoftmargin} for binary classification. 
 \vspace{-0.35cm}
\subsection{Density Branch}
\label{density}
 The image classification branch described above predicts the presence or absence of objects  using the class confidence scores derived from the peak map $\tilde {\textbf{M}}^{c}$.  The object category map obtained from the image classifier branch provides a coarse localization of objects, which can provide spatial guidance during the training of our density branch. 
 However, it struggles to differentiate between multiple objects and single-object parts due to the lack of prior information about the number of object instances (see Fig.~\ref{fig:SegmentIntro}(b)). This causes a large number of false positives in the peak map $\tilde{\textbf{M}}^{c}$.   To address this issue, we introduce a hard spatial-guidance module that utilizes the count information and generates a spatial  mask from the peak map $\tilde{\textbf{M}}^{c}$.

      \subsubsection{Hard Spatial-guidance}
      \label{sec:hardSpatialGuidance}
      Here,  the coarse-localization  ability of the object category map  is used to  generate a spatial  mask.  
  For all object categories $c \in S$, the ${t_c}$-{th} highest peak value of peak map $\tilde{M}^{c}$ is computed using the heap-max algorithm \cite{max_heap}. The ${t_c}$-{th} highest peak value $h_{c}$ is then used to generate a spatial  mask $\textbf{B}^{c}$ as,

 \vspace{-0.1cm}
\begin{equation} 
    \textbf{B}^{c}=u(\tilde{\textbf{M}}^{c}-h_c).
     \vspace{-0.1cm}
\end{equation}
 Here, $u(n)$ is the unit step function which is $1$ only if $n \ge 0$.   We use the spatial  mask as a pseudo-ground truth mask to compute the spatial loss term used for training the density branch.  This supports the preservation of the spatial distribution of object counts in a density map. Later, we show that this property helps to improve instance segmentation (Sec.~\ref{sec_instance_seg}).

      \subsubsection{Category-specific Density Map}
            \label{sec:categoryspecificdensityMap}
      The density branch produces a category-specific density map $\textbf{D}^c$, where  each pixel  indicates the number of objects belonging to  category~$c$  in the corresponding image region. Additionally, the accumulation of $\textbf{D}^c$  over any image region  estimates the instance count of category~$c$ in that region \cite{OxfordDensityNIPS2010}.  On the other hand, the pixels in the object category map  $\textbf{M}^c$  indicate the confidence that the corresponding image pixels belong to  object category~$c$.

   When constructing a density map, it is desired to estimate accurate object counts in every image sub-region. Our spatial loss term ${\cal{L}}_{spatial}$ in Eq.~\ref{eq:maineq} ensures that individual object instances are localized while the count term ${\cal{L}}_{count}$ constrains the category-specific object count to that of the ground-truth count. These terms are explained next. 
  
  \noindent\textbf{Spatial Loss:}  
The spatial loss  ${\cal{L}}_{spatial}$ is divided into the loss $\mathcal{L}_{sp+}$ which enhances the positive peaks corresponding to instances of object categories within ${S}$, and the loss $\mathcal{L}_{sp-}$ which suppresses false positives of categories within ${S_0}$.   
 Due to the unavailability of absolute object counts, the set $\tilde{S}$ is not used in the spatial loss and treated separately later. 
  To enable LC supervised density map training using  ${\cal{L}}_{spatial}$, we employ the spatial mask $\textbf{B}^{c}$ as a pseudo ground-truth mask. 
 
 Although the non-zero elements of the spatial mask $\textbf{B}^{c}$ indicate object locations, its zero elements do not necessarily point towards the background. Therefore, we construct a masked density map $\tilde{\textbf{D}}^{c}$ to exclude density map $\textbf{D}^{c}$ values at locations where the corresponding $\textbf{B}^{c}$ values are zero. Those density map $\textbf{D}^{c}$ values should also be excluded during the loss computation in Eq.~\ref{eq:possptial} and backpropagation, due to their risk of introducing false negatives. This is achieved by computing the Hadamard product between the density map $\textbf{D}^{c}$ and $\textbf{B}^{c}$ as,
 \vspace{-0.1cm}
\begin{equation} 
\label{eq:hadmard}
    \tilde{\textbf{D}}^{c}=\textbf{D}^{c} \odot \textbf{B}^{c}.
    \vspace{-0.1cm}
\end{equation}
The spatial loss $\mathcal{L}_{sp+}$ for object categories within the lower-count range ${S}$ is computed between $\textbf{B}^{c}$ and $\tilde{\textbf{D}}^{c}$ using a logistic binary cross entropy (logistic BCE) \cite{pytorch_cite} loss for positive ground-truth labels. 
 The logistic BCE loss transfers the network prediction ($\tilde{\textbf{D}}^{c}$) through a  sigmoid activation  layer $\sigma$ and computes the standard BCE loss as,

 \vspace{-0.1cm}
   \begin{equation} 
    \label{eq:possptial}
  \mathcal{L}_{sp+}(\tilde{\textbf{D}}^{c},~\textbf{B}^{c}) = - \sum_{\forall c \in S} \frac{\texttt{sum}(\textbf{B}^{c} \odot\log ( \sigma(\tilde{\textbf{D}}^{c})) )} {|S|\cdot \texttt{sum}(\mathbf{B}^{c})}.
   \vspace{-0.1cm}
\end{equation} 
Here, $|S|$ is the cardinality of the set $S$ and $\texttt{sum}(\cdot)$ is computed by taking the summation over all elements in a matrix. For example,  $\texttt{sum}(\mathbf{B}^{c})$ = $\mathbf{1}^h \mathbf{B}^{c} \mathbf{1}^w$, where $\mathbf{1}^h$ and $\mathbf{1}^w $ are all-ones vectors of size ${1\times H}$  and ${W\times 1}$, respectively.
Here, the highest $t_c$ peaks in $\tilde{\textbf{M}}^{c}$ are assumed to fall on $t_c$  instances of object category $c \in S$.
Due to the unavailability of ground-truth object locations, we use this assumption and observe that it holds in most scenarios.

The spatial loss $\mathcal{L}_{sp+}$ for the positive ground-truth labels enhances positive peaks corresponding to instances of object categories within ${S}$. However, the false positives of the density map for $c \in S$ are not penalized in this loss. We therefore introduce another term,  $\mathcal{L}_{sp-}$, into the loss function to address the false positives of $c\in S_0$.

For  $c\in S_0$,  positive  activations of $\textbf{D}^{c}$ indicate false detections. A zero-valued mask ${\mathbf{0_{H\times W}}}$ is used as ground-truth to reduce such false detections using the logistic BCE loss,
 \vspace{-0.1cm}
    \begin{equation} 
  \mathcal{L}_{sp-}(\mathbf{D}^c,\mathbf{0}_{H\times W}) = -\sum_{c\in S_0}{{ \frac{\texttt{sum}(\log (1- \sigma({\textbf{D}}^{c}))) }{|S_0|\cdot H\cdot W}}}.
   \vspace{-0.1cm}
\end{equation}
Though the spatial loss ensures the preservation of the spatial distribution of objects, only relying on local information may result in deviations in   object count. 

\noindent\textbf{Count Loss:}
The count loss penalizes deviations of the predicted count $\hat{t_c}$ from the ground-truth. It has two components: the ranking loss $\mathcal{L}_{rank}$ for object categories beyond the lower-count range (i.e., $\forall c\in \tilde{S}$) and the mean-squared error (MSE) loss $\mathcal{L}_{MSE}$ for the rest of the categories.  
$\mathcal{L}_{MSE}$ penalizes the predicted density map if the category-specific count prediction does not match with the ground-truth count,  i.e., 
 \vspace{-0.1cm}
\begin{equation} 
\mathcal{L}_{MSE}(\hat{t_c},t_c)= \sum_{c\in \{S_0,S\}}\frac{({\hat{t_c}-t_c)^2 }}{\small{|S_0|+|S|}}.
\label{eq:loss_mse}
 \end{equation}
 Here, the predicted count $\hat{t_c}$ is the  accumulation of the density map for a category $c$ over its entire spatial region,  \ie,   
  $\hat{t_c}=\texttt{sum}(\textbf{D}^{c})$. 
Note that object categories in $\tilde{S}$ were not previously considered when computing the spatial loss $\mathcal{L}_{spatial}$ or mean-squared error loss $\mathcal{L}_{MSE}$. Here, we introduce a ranking loss \cite{rankingloss_cvpr2014} with a zero margin that penalizes under-counting for object categories within $\tilde{S}$,

\begin{equation} 
\mathcal{L}_{rank}(\hat{t_c}, \tilde{t})=\sum_{c \in \tilde{S}}{\!{\frac{ \max(0, \tilde{t}-{\hat{t_c}})}{|\tilde{S}|}}}.
\label{eq:loss_rank}
 \end{equation}
The ranking loss penalizes the density branch if the predicted object count $\hat{t_c}$ is less than $\tilde{t}$ for $c\in \tilde{S}$. Recall, the beyond lower-count range $\tilde{S}$ starts from $\tilde{t}=5$.

Within the lower-count range $S$, the spatial loss term $\mathcal{L}_{spatial}$ is optimized to locate object instances,  while the  MSE loss ($\mathcal{L}_{MSE}$) is optimized to accurately predict the corresponding category-specific count. Due to the joint optimization of both these terms within the lower-count range, the network learns to correlate between the located objects and the object count. Further, the network is able to locate object instances, generalizing beyond the lower-count range $\tilde{S}$ (see Fig.~\ref{fig:SegmentIntro}). Additionally, the ranking loss $\mathcal{L}_{rank}$ term in the proposed loss function ensures the penalization of  under-counting for cases beyond the lower-count  range $\tilde{S}$.

\noindent\textbf{Mini-batch Loss:} The normalized loss terms $\mathcal{\hat{L}}_{sp+}$, $\mathcal{\hat{L}}_{sp-}$, $\mathcal{\hat{L}}_{MSE}$ and $\mathcal{\hat{L}}_{rank}$ are computed by averaging their respective loss terms over  all  images in the mini-batch. Then, the  average spatial loss over a mini-batch,    
$\mathcal{L}_{spatial}$, is computed by $\mathcal{\hat{L}}_{sp+}\llbracket c\in S \rrbracket +  \mathcal{\hat{L}}_{sp-}\llbracket c\in {S_0} \rrbracket$, where $\llbracket \cdot \rrbracket$ denotes Iverson brackets.   For categories beyond the lower-count range,  $\mathcal{\hat{L}}_{rank}$ can lead to over-estimation of the count. Hence, the  overall count loss $\mathcal{L}_{count}$ is computed by  assigning  a relatively low weight ($\lambda=0.1$) to $\mathcal{\hat{L}}_{rank}$ (see Table~\ref{tab:loss_analysis}), \ie,  $ \mathcal{L}_{count}=\mathcal{\hat{L}}_{MSE}\llbracket c\in {S_0, S} \rrbracket+  \lambda* \mathcal{\hat{L}}_{rank}\llbracket c\in \tilde{S} \rrbracket$.

\HMC{Both branches of our architecture are trained simultaneously using a two stage training strategy (see Eq.~\ref{eq:maineq}).  In the first stage, the density branch is trained by excluding the spatial loss ${\cal{L}}_{spatial}$  that requires pseudo ground-truth generated from the image classification branch. The second stage includes the spatial loss for training the density branch. In both stages, the image classification branch is trained using the multi-label image-classification loss ${\mathcal{L}}_{class}$.}\\

\noindent\textbf{Backpropagation:}
\label{sec:backpropagate}
   \HMC{As mentioned earlier, our density branch training  exploits the localization capability of image classification branch to obtain  a density map that follows the object locations captured by the object category map. 
  We use the spatial mask $\textbf{B}^{c}$ derived from the image classification branch as a pseudo ground-truth  to train the density branch.   
  Since the networks trained with counting (regression) objective may have inferior localization capability compared to the classification branch,  back-propagation of the gradients through $\textbf{B}^{c}$ to the image-classification branch,  based on the  spatial loss in the density branch may result in dislocating  peaks  in the object category map, and thereby adversely effecting the training of density map in successive epochs.  Hence, we use the spatial mask $\textbf{B}^{c}$ as a pseudo ground-truth avoiding backpropagation (shown with green arrows in Fig.~\ref{Fig:architectue}),  which further helps to obtain a stable image classification branch, and a density map that follows the object locations captured by the object category map. }

The image classification branch is backpropagated as in \cite{PRM}. 
 In the density branch,  we use the Hadamard product of the density map with  $\textbf{B}^{c}$ in Eq.~\ref{eq:hadmard} to compute $\mathcal{L}_{sp+}$ for $c\in S $.  Hence, the  gradients ($\delta^{c}$) for the $c^{th}$ channel of the last convolution layer of the density branch, due to $\mathcal{L}_{sp+}$, are  computed as, 
  \vspace{-0.1cm}
\begin{equation} 
    \delta_{sp+}^c= \pd{\mathcal{\hat{L}}_{sp+}}{\tilde{\textbf{D}}^{c} } \odot \textbf{B}^{c}. 
     \vspace{-0.1cm}
\end{equation}
Since   $\mathcal{L}_{MSE}$, $\mathcal{L}_{rank}$ and $\mathcal{L}_{sp-}$   are computed using  MSE, ranking and logistic BCE losses on convolution outputs, their respective gradients  are computed using off-the-shelf Pytorch implementation \cite{pytorch_cite}.  
Next, we describe the extension of our LC framework to enable object counting in a reduced lower-count (RLC)  supervised setting. 
 
    	\begin{figure*}[t]
		\centering
	\includegraphics[width=1.0\linewidth, clip=true, trim=0cm 0cm 0cm 40cm]{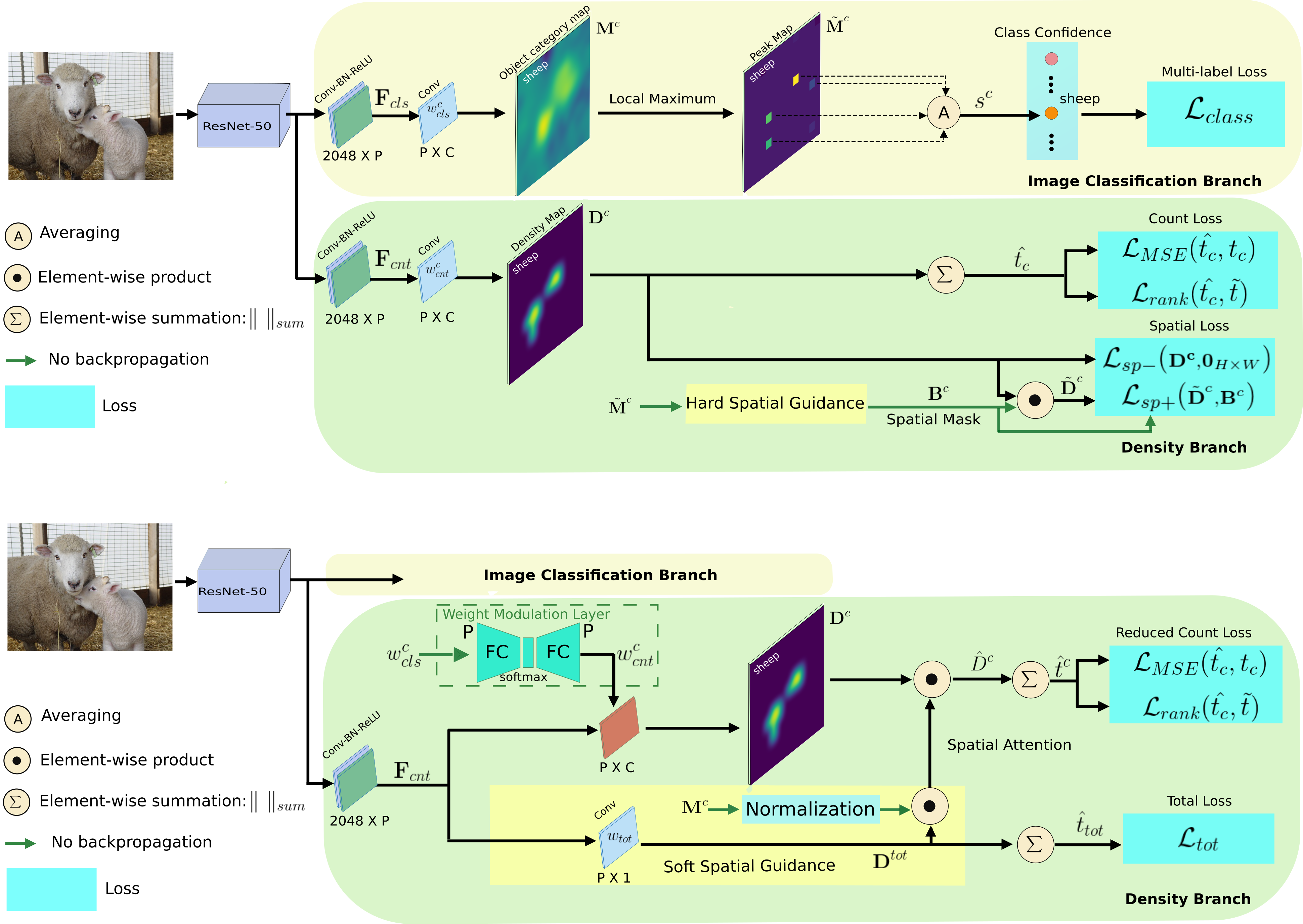}\\ 
			\caption{Overview of our RLC architecture, which comprises  an image classification branch and a density branch. The image classification branch has an identical structure as the  LC  architecture, and is trained on all categories ($\mathcal{A} \cup \mathcal{B}$) using the class labels indicating the presence or absence of the objects. 	The density branch has two sub-branches: a category-specific and a  category-independent density sub-branch.  The category-specific sub-branch adapts the convolution weights from the image classifier branch using a weight modulation layer and then generates a category-specific map, \HMC{which is  
			multiplied with  a spatial attention map 
			 to obtain the category-specific density map  ($\hat{D}^c$) and  category-specific counts.}    
	Training  this branch updates the weight modulation layer,   and only categories  with known counts (set $\mathcal{A}$) are used for the training. 
	The convolution operator using adapted convolution weights is shown in red color. The category-independent density sub-branch predicts the total counts of all objects. }
				\label{fig:SPC_archit}
			\vspace*{-0.2cm}
\end{figure*}

 \section{Reduced Lower-count Supervised Framework}
 \label{SPC_overview}
 The LC supervision requires counts within the lower-count range for `\emph{all}' object categories. These are  laborious to obtain for natural scene datasets that have a large number of object categories (\eg,  Visual Genome).  Here, we introduce an even more  challenging setting, named  reduced lower-count (RLC) supervision, which further reduces the level of supervision  such that counts within the lower-count range are known for only a subset of categories during training. For this setting, we propose an RLC framework that extends our dual-branch LC framework and predicts counts for \textit{all categories}, irrespective of the count annotations.

   In our RLC supervised setting, the set $\mathcal{A}$ indicates the object categories in the \textit{dataset} for which both the category labels and count annotations are known, whereas  the set $\mathcal{B}$ indicates the categories whose category labels are known, but count annotations
  are not available. We employ a LC annotation  to the object categories in $\mathcal{A}$. This further splits $\mathcal{A}$ into three disjoint subsets: $S_0$,   $S$ and $\tilde{S}$, based on the \textit{object count in image} $I$ (see Table~\ref{table:notations}).   As mentioned in Sec.~\ref{subsec:ProposedLossFn},  the set $S_0$ indicates the object categories which are absent in image $I$ (\ie, $t_c$=0). Similarly, the set $S$   represents the categories within the lower-count  range ($t_c < \tilde{t}$), where $\tilde{t}=5$. 
  Finally, the set $\tilde{S}$  indicates the categories beyond the lower-count range ($t_c \ge \tilde{t}$) whose \textit{exact} count annotations are not available. This RLC supervised setting is highly challenging compared to an LC supervised setting (Sec.~\ref{proposedmethod_start}) that has  $\mathcal{B}=\emptyset$.

 \noindent\textbf{The Proposed RLC Architecture:}
 Similar to the LC architecture, our RLC architecture has an image classification branch and a density branch   built on a ResNet50 backbone (See Fig.~\ref{fig:SPC_archit}). 
 Here, we explain the changes introduced to our LC architecture (Fig.~\ref{Fig:architectue}) for count prediction in the challenging RLC setting.  First, we introduce a novel weight modulation layer (Sec.~\ref{subsec:weightModulationLayer}) that transfers knowledge from object categories with annotated counts to the categories where count information is unknown. We then introduce an auxiliary category-independent sub-branch to the density branch to produce a category-independent density map $\mathbf{D}^{tot}$ (Sec.~\ref{Sec:softSpatialGuidance}). This additionally introduced category-independent density sub-branch  provides category-independent total count predictions for an image, including the object categories without count annotations.  In summary, the density branch in our RLC framework has two sub-branches; namely, the category-specific sub-branch and category-independent sub-branch, both of which  share a common input $\mathbf{F_{cnt}}$, as shown in Fig.~\ref{fig:SPC_archit}.

  The object category  maps (class activation maps with boosted local maxima) used in the image classification  branch can provide a coarse localization of object instances from their respective categories.  With LC supervision,   this coarse-localization ability  of the object category map can be used along with the category-specific LC count information, to generate a spatial mask (see Sec.~\ref{sec:hardSpatialGuidance}) that preserves the spatial information for the  category-specific density map training. However, in RLC supervision, category-specific count annotations are not available for a set of object categories (the set $\mathcal{B}$). Therefore, employing such a hard spatial-guidance strategy is infeasible for obtaining a spatial mask, to train the category-specific density map.  To alleviate this limitation, we introduce a soft spatial-guidance strategy that uses the category-independent density map $\mathbf{D}^{tot}$  along with the object category  map $\textbf{M}^c$ of the image classifier to produce a  spatial attention map $\mathbf{G}^c$ for the  category~$c$. This spatial attention map is used to provide  spatial information to the category-specific density map.  We train our RLC framework with a novel loss function described below.

\noindent\textbf{The Proposed Loss Function:} The complete network, comprising of an image classification and a  density branch, is jointly trained in an end-to-end fashion using the proposed objective, given below:  
\begin{equation}
    \mathcal{L}=\mathcal{L}_{cls}+\mathcal{L}_{rcount}+\mathcal{L}_{tot},
\end{equation}
where $\mathcal{L}_{cls}$ refers to the multi-label classification loss used to train the classification branch.  The second and third terms are used to train the density branch, where   $\mathcal{L}_{rcount}$ is used to train the category-specific  sub-branch, and  $\mathcal{L}_{tot}$ is used to train the category-independent one.

Since  category labels are available for all object categories, the \textit{image classifier branch}  is trained on all object categories using  $\mathcal{L}_{cls}$, without introducing any changes compared to Sec.~\ref{peakstimulation}. 
 Considering that  count annotations are only available  for  set $\mathcal{A}$, we adapt the   density branch for  RLC supervision,  as explained in the next section.

\subsection{Density Branch}\label{subsec:catspec} 
To address the unavailability of count annotations for  set $\mathcal{B}$ in our RLC setting,   we introduce a weight modulation layer that adapts the convolution weights from the image classifier branch to generate category-specific density maps that estimate category-specific object counts for \textit{all} categories. In this section, we first introduce this weight modulation layer, followed  by the generation of category-specific  density maps with  RLC supervision. 

\subsubsection{Weight Modulation Layer}
\label{subsec:weightModulationLayer}

The weight modulation layer $\Psi$  is class-agnostic by nature and is learned using the object categories that have  both class and count annotations (set $\mathcal{A}$). It is then used to compute  the convolution weights of the category-specific  sub-branch  for all categories, including those categories  without count annotations (set $\mathcal{B}$). {Since object counting looks for repetitive patterns in a category-specific projected embedding space, the function $\Psi$ learns a generalizable mapping that works alike for objects with known and unknown count annotations (RLC  supervised setting)}.

Let $\bm{w}^c_{cls}$ be the convolution weights in the image classification branch (trained using the set $ \mathcal{A}\cup \mathcal{B}$)  for the object category $c$, and $\mathbf{D}=\{\mathbf{D}^c {\in} \mathbb{R}^{H\times W}: c{\in}[1,C]\}$ represent the category-specific  maps
\HMC{used for obtaining the category-specific density maps $\hat{\mathbf{D}}^c$  
for all $C$ object categories.} The weights $\bm{w}^c_{cls}$ are passed through the weight modulation layer $\Psi$, resulting in the convolution weights $\bm{w}^c_{cnt}$  that produce  $\mathbf{D}^c$  
for   category $c$, \ie, 

 \begin{equation}
    \bm{w}^c_{cnt}=\Psi(\bm{w}^c_{cls}) , \; \text{and~} ~\mathbf{D}^c = \bm{w}^c_{cnt} * {\mathbf{F_{cnt}}}, 
 \end{equation}
 where `$*$' denotes convolution.  Note that this convolution operator (shown  in red  in Fig.~\ref{fig:SPC_archit}) has no learnable parameters. Instead, it uses the $\bm{w}^c_{cnt}$ obtained from the $\bm{w}^c_{cls}$ through the weight modulation layer. 

 The proposed weight modulation layer is shown in Fig.~\ref{fig:SPC_archit}. It passes $\bm{w}^c_{cls}$  through a $P\times \frac{P}{2}$ fully connected layer, followed by a softmax non-linear layer for normalization, and finally through another $\frac{P}{2} \times P$ fully connected layer.  
 The low-dimensional embedding (to $\frac{P}{2}$)  by the first fully connected layer acts as a bottleneck architecture that leads to superior performance. We conjecture that the bottleneck architecture of the modulation layer $\Psi$ projects the convolution weights from  similar  object categories to similar representations, which enables  generalization of the weight modulation layer to the categories whose count annotations are not available. Backpropagation is not performed   
 from $\Psi$ to the image classifier branch (shown with green colored  line in Fig.~\ref{fig:SPC_archit}) to avoid creating a discrepancy in $\bm{w}^c_{cls}$ between $\mathcal{A}$ and $\mathcal{B}$, since only $\bm{w}^c_{cls}$ for $c \in \mathcal{A}$ will receive gradient signals from $\Psi$ while the image classifier ($\bm{w}^c_{cls}$) is trained on both $\mathcal{A}$ and $\mathcal{B}$. 

\subsubsection{Soft Spatial-guidance}
\label{Sec:softSpatialGuidance} 
 The object category maps (class activation maps with boosted local maxima)   used in the classification branch can provide a coarse localization of object instances from their respective categories.  In our LC setting (Sec.~\ref{sec:hardSpatialGuidance}),  this coarse-localization ability  of object category maps is used along with the category-specific count information to generate a spatial mask that preserves the spatial information for the  density map training. Category-specific count information is not available for  set $\mathcal{B}$, so such a  spatial mask cannot be used in the RLC framework.  
 To alleviate this limitation, the category-independent density map $\mathbf{D}^{tot}$ is used along with the object category map $\textbf{M}^c$ of the image classifier to produce a spatial attention map $\mathbf{G}^c$ for   category~$c$.  
 Considering the large variations in ${\mathbf{M}}^c$, we pass it through a sigmoid activation and obtain a normalized map $\hat{\textbf{M}}^c$.  The spatial attention for category $c$,  $\mathbf{G}^c$, is computed as,

  \begin{equation}
  \mathbf{G}^c=\hat{\textbf{M}}^c\circ \mathbf{D}^{tot}.
  \end{equation}
The spatial attention map $\mathbf{G}^c$ thus obtained is used  with the  corresponding  $\mathbf{D}^c$ to obtain the category-specific density map $\hat{\mathbf{D}}^c$. Next, we introduce the procedure to generate a category-independent density map that provides total object counts for $\mathbf{I}$.

\noindent{\textbf{Total-count Loss and  Category-independent  Density  Map: }}\label{subsec:catind}
 The category-independent density map $\mathbf{D}^{tot}$  is generated using a category-independent sub-branch (operating on input feature $\mathbf{F_{cnt}}$)   that consist of an additional $1\times1$ convolution with a single output channel. 
 The category-independent total count of objects in an image can be estimated by accumulating  $\mathbf{D}^{tot}$ over its entire spatial region, \ie,   $\hat{t}_{tot}=\sum_{ij}{\text{D}}^{tot}_{ij}$.  We use only the count-annotated object categories $S$ and $\tilde{S}$ to train the category-independent density branch, and  the  ground-truth  total count $t_{tot}$ for the image $I$ is computed  as  
\begin{equation}
    t_{tot}=\sum_{c\in S} t_c+ (\tilde{t}\times |\tilde{S}|).
\end{equation}
The category-independent density branch is trained with the following loss function, 
 \begin{equation}
  \mathcal{L}_{tot}= \mathcal{L}_{MSE}(\hat{t}_{tot}, t_{tot}) \llbracket Z=0 \rrbracket +    \mathcal{L}_{rank}(\hat{t}_{tot}, t_{tot}) \llbracket Z > 0\rrbracket.
\end{equation}
 $Z$ is the total number of object categories  whose exact count is unknown, i.e., $Z{=} |\tilde{S}|{+}|\mathcal{B}'|$, where $\mathcal{B}'$ is the set of positive categories of $\mathcal{B}$ in image $I$.
 Here,   $\mathcal{L}_{MSE}(\hat{t}_{tot}, {t}_{tot})=(\hat{t}_{tot}-t_{tot})^2$ is the mean-squared error loss and $\mathcal{L}_{rank}(\hat{t}_{tot}, t_{tot})$ is the ranking loss, such that:
  \begin{equation}
  \mathcal{L}_{rank}(\hat{t}_{tot}, t_{tot})= \max(0,t_{tot}-\hat{t}_{tot}).
 \end{equation}
 
Next, we describe the generation of category-specific density map. 

 \subsubsection{Category-specific Density Map}
 As mentioned in Sec.~\ref{subsec:weightModulationLayer}, the category-specific  map  $\mathbf{D}^c$ for class $c$ is generated using  the  modulated convolution weights $\bm{w}^c_{cnt}$ (obtained using $\Psi$). 
We use a category-specific 
spatial attention  $\mathbf{G}^c$  to preserve the spatial information in the density map  $\hat{\mathbf{D}}^c$, as
     \begin{align}
   \hat{\mathbf{D}}^c=\mathbf{D}^c \circ \mathbf{G}^c,
  \end{align}
  where `$\circ$' denotes the Hadamard product. 
  The  density map $\hat{\mathbf{D}}^c$ is used to estimate the  category-specific object count $\hat{t}_c$ by accumulating $\hat{\mathbf{D}}^c$  over its entire spatial region, \ie, $\hat{t}_c=\sum_{ij}\hat{\text{D}}_{ij}^c$, where $i,j$ are indices of matrix elements.
  
  The category-specific  sub-branch is trained with the \textit{reduced-count loss function} $\mathcal{L}_{rcount}$, 
 \begin{equation}
   \mathcal{L}_{rcount}=
   \mathcal{L}_{MSE}(\hat{t}_{c},~t_{c}) \llbracket c\in S \rrbracket + 
    \mathcal{L}_{rank}(\hat{t}_{c},~\tilde{t}) \llbracket c\in \tilde{S}\rrbracket.
    \label{eq:L_cat}
\end{equation}
 Here,  $\mathcal{L}_{MSE}$  and $\mathcal{L}_{rank}$ are computed using Eq.~\ref{eq:loss_mse} and Eq.~\ref{eq:loss_rank}, respectively. It is worth noting that the predicted count  $\hat{t}_{c}$ is obtained from a spatial attention ($\mathbf{G}^c$) weighted  density map $\hat{\mathbf{D}}^c$. Hence, the spatial distribution of objects is preserved to a  certain degree while  minimizing the count error in  Eq.~\ref{eq:L_cat} (see Fig.~\ref{fig:densitymapsLC_RLC}). Considering that the modulated convolution weight $\bm{w}^c_{cnt}$ used to generate   $\mathbf{D}^c$  is obtained through $\Psi(\bm{w}^c_{cls})$, minimizing  $\mathcal{L}_{rcount}$ will result in training   the class-agnostic weight modulation layer $\Psi$.

\section{Training and Inference}

Throughout our experiments, we use a fixed set of training hyper-parameters. A ResNet-50 backbone architecture  pre-trained on the ImageNet dataset is used in our evaluations. The backbone network is trained with an initial learning rate of $10^{-4}$, while the image classification and  density branches are trained with an initial learning rate of $0.01$. The number of input channels $P$ of the final $1{\times}1$ convolutions in  each branch is empirically set to $P=1.5{\times}C$ across datasets, based on  experiments performed on the validation sets.
A mini-batch size of $16$ is used for the SGD optimizer. The momentum and weight decay are set to  0.9 and $10^{-4}$, respectively. Our algorithm is implemented using Pytorch \cite{paszke2017automatic} on a Tesla V100 GPU. Our code will be made publicly available for reproducible research. 

 \noindent\textbf{Inference:} 
 For both LC and RLC supervision, 
the image classification branch outputs a class confidence score $s^c$ for each class, indicating the  presence (~$\hat{t_c}>0$, if $s^c>0$) or absence ($\hat{t_c}=0$, if  $s^c \le 0$ ) of   object category $c$.  
With LC, the predicted count $\hat{t_c}$ is obtained by summing the density map $\textbf{D}^c$ for  category $c$ over its entire spatial region, whereas with RLC supervision,  $\hat{t_c}$ is obtained from the  density map $\hat{\mathbf{D}}^c$.  The proposed approach only utilizes   count annotations within the lower-count range ($t_c\le 4$) and accurately predicts object counts for \emph{both} within and beyond the  lower-count range  (see Fig.~\ref{fig:ablation_subitizingRange} and Fig.~\ref{Fig:qual_coco}).  Additionally, the accumulation of the category-independent density map  in our RLC architecture leads to a total count for  all objects, irrespective of their category (Sec.~\ref{subsec:catind}). 

\vspace{-0.1cm}
\section{Experiments}

\noindent\textbf{Datasets:}
We evaluate our method on three challenging datasets: PASCAL VOC 2007 \cite{pascal_2012},  COCO \cite{coco_eccv2014} and  Visual Genome \cite{krishnavisualgenome}.
For fair comparison with existing approaches on PASCAL VOC 2007 \cite{pascal_2012} and  COCO, we employ the same splits, named as count-train, count-val and count-test, as used in the state-of-the-art methods \cite{WhereAreBlobsECCV18,Chattopadhyay_2017_CVPR}.

Specifically, for the PASCAL VOC dataset, the training, validation and test set are used as count-train, count-val and count-test, respectively. For the COCO dataset, the training set is used as count-train, the first half of the validation set as the count-val, and the second half as  count-test.  
The best models on the count-val set are used to report the results  on the  count-test set.

Our RLC framework aims at reducing the annotation cost for  applications targeting large numbers of object categories. Hence, we only use COCO and Visual Genome (VG) for evaluation since these datasets contain a large number of object categories. The COCO dataset has 80 object categories and Visual Genome (VG) dataset has 80,000 classes. 
Following \cite{zero_shot_eccv18}, we remove non-visual classes from the VG dataset and consider the remaining 609 classes.

To evaluate image-level supervised instance segmentation \cite{PRM, Zhu_2019_CVPR,laradji2019masks,IRnet},  we train and report the results on the PASCAL VOC 2012 dataset, similar to \cite{PRM}. Specifically, our model is trained on an augmented set of 10,582  training images, provided by \cite{pascal_context},   using only image-level lower-count supervision.  The performance is evaluated on 1,449 object segmentation images from the validation set  of PASCAL VOC 2012. 
 
\begin{table}[t]
\resizebox{\columnwidth}{!}{
\begin{tabular}{>{\centering\arraybackslash}p{3cm}|c|cccc}
\hline 
 Approach &SV&mRMSE&  \begin{tabular}[c]{@{}c@{}}mRMSE\\\vspace{-0.1cm} -nz\end{tabular}   & \begin{tabular}[c]{@{}c@{}} m-rel\\RMSE\end{tabular}   &\begin{tabular}[c]{@{}c@{}} m-rel\\RMSE-nz \end{tabular} \\\hline
CAM+MSE& IC&	\sbest{0.45}&	\sbest{1.52}&	\sbest{0.29}&	\sbest{0.64}\\\hline
Peak+MSE&IC	&0.64	&2.51&	0.30&	1.06\\\hline  \hline
Proposed (LC)  & LC & \best{0.29} & \best{1.14} & \best{0.17} & \best{0.61}\\\hline
\end{tabular}}\vspace{0.2em}
\caption{Counting performance on the Pascal VOC 2007 count-test set using our approach and two baselines. Both baselines are obtained by training the network using the MSE loss function.} 
\label{tab:counting_pascal_baseline}
\vspace{-0.35cm}
\end{table}
\vspace{-0.15cm}
\begin{figure}[t]
		\centering
						\includegraphics[width=0.98\linewidth, clip=true, trim=0cm 15cm 11.5cm 0cm]{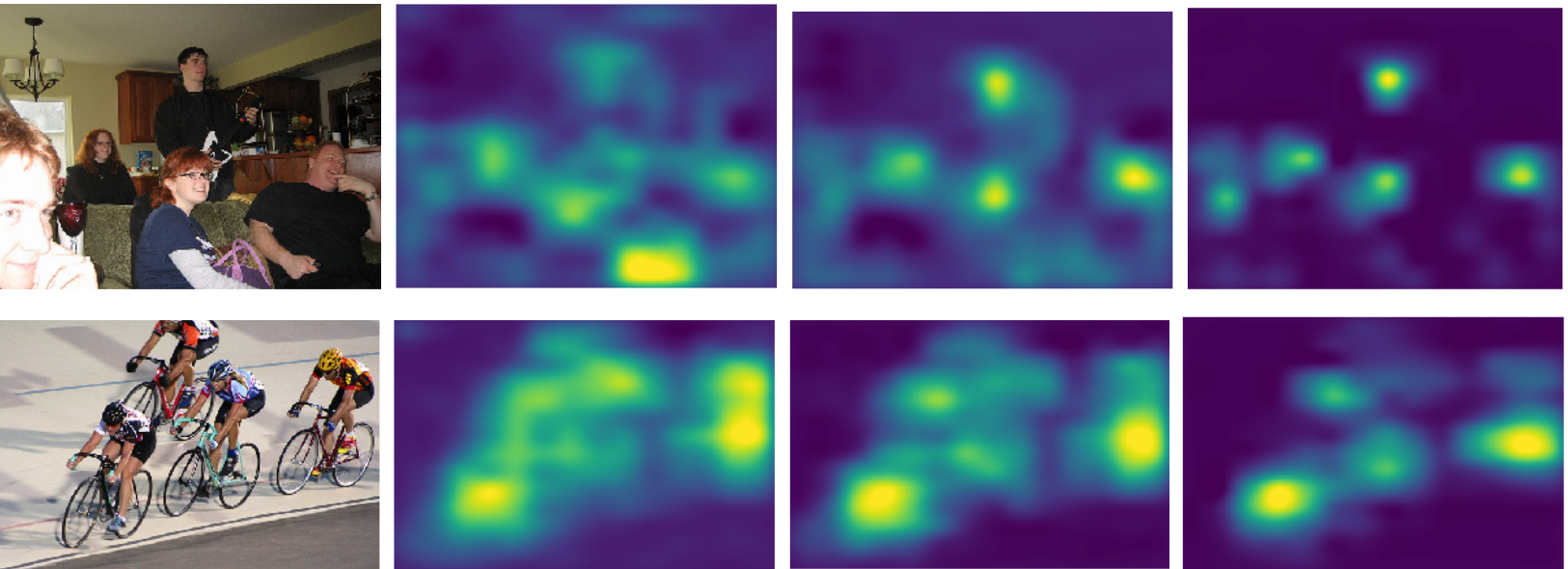}\\ \vspace{-0.1cm}
						\resizebox{\linewidth}{!}{\hspace*{0.06\linewidth} (a) Input Image \hspace*{0.1\linewidth} (b) Class+MSE \hspace*{0.16\linewidth}(c) +Spatial \hspace*{0.15\linewidth} (d) +Ranking  \hspace*{0.08\linewidth} }\\
				\vspace*{-0.0cm}
			\caption{Progressive improvement in density map quality with the incremental introduction of spatial and ranking loss terms in our LC framework. In both cases (top row: person and bottom row: bicycle), our overall loss function integrating all three terms provides the best density maps. The category-specific object count is accurately predicted (top row: 5 persons and bottom row: 4 bicycles) by accumulating the respective density map.   
				}
			\label{Fig:experi_densitymap}
			\vspace*{-0.3cm}
\end{figure}
 \noindent\textbf{Evaluation Criteria:}
 Following previous works\cite{Chattopadhyay_2017_CVPR, WhereAreBlobsECCV18}, we evaluate generic object counting using the root-mean-square error (RMSE) metric and its three variants; namely, RMSE non-zero (RMSE-nz), relative RMSE (relRMSE) and relative RMSE non-zero (relRMSE-nz). The $RMSE_c$ and  $relRMSE_c$ errors for a category $c$ are computed as $\sqrt{\frac{1}{T}\sum_{i=1}^{T}(t_{ic}-\hat{t_{ic}})^2}$ and   $\sqrt{{\frac{1}{T}\sum_{i=1}^{T}\frac{(t_{ic}-\hat{t_{ic}})^2}{t_{ic}+1}}}$,  respectively. Here, $T$  is the total number of images in the test set and $\hat{t}_{ic}$, $t_{ic}$ are the predicted count (rounded to the nearest integer) and the ground-truth count for category $c$ in an image $i$.  
 The errors are then averaged across all categories to obtain the mRMSE and m-relRMSE on a dataset. The above metrics are also evaluated for ground-truth instances with non-zero counts as mRMSE-nz and m-relRMSE-nz. To evaluate the category-independent total count $\hat{t}_{tot}$, we only use RMSE and relRMSE since each image has at least one instance and the RMSE-nz metric is equal to the RMSE metric. For all error metrics mentioned above, \textit{smaller} numbers indicate \textit{better} performance. We refer to  \cite{Chattopadhyay_2017_CVPR} for more details. For instance segmentation, the performance is evaluated using Average Best Overlap (ABO) \cite{ABO_2015} and ${mAP}^r$, as in \cite{PRM}. The ${mAP}^r$ is computed with intersection over union (IoU) thresholds of 0.25, 0.5 and 0.75. Throughout our experiments (Table  \ref{tab:counting_pascal_baseline} - \ref{tab:ins_seg_pascal}) , we report the top-two results in red and blue font, respectively. 

 \noindent\textbf{Supervision Levels:} The level of supervision is indicated as SV in Tables  \ref{tab:counting_pascal_baseline},  \ref{tab:counting_coco}, \ref{tab:results_vg1} and \ref{tab:counting_pascal}. BB indicates bounding box supervision and PL indicates point-level supervision for each object instance. Image-level supervised methods using only within subitizing or lower-count range counts are denoted as LC, while the methods using both within and beyond subitizing range counts are indicated as IC.   Method using reduced lower-count supervison is indicated as RLC.

\subsection{Ablation Study}
\label{ILC_exp}

\label{expe:everyday object counting results}
\subsubsection{Counting with Lower-count Supervision (LC)}
\noindent\textbf{Importance of Dual-branch Architecture: }
We perform an ablation study on the PASCAL VOC 2007 count-test. First, the impact of our dual-branch architecture is analyzed by comparing it with two baselines: class-activation \cite{CAM} based regression (CAM+MSE) and peak-based regression (Peak+MSE), using the local-maximum boosting approach of \cite{PRM}. Both baselines are obtained by end-to-end training of the network, employing the same backbone, and using the MSE loss function to directly predict the category-specific object count. Table \ref{tab:counting_pascal_baseline} shows the comparison. Our approach largely outperforms both baselines, highlighting the importance of having a dual-branch architecture. 
Next, we evaluate the contribution of each term in our loss function towards the final counting performance.

\noindent\textbf{Importance of Different Loss terms: }
 Fig.~\ref{Fig:experi_densitymap} shows the systematic improvement in the quality of density maps (top row: person and bottom row: bicycle) with the incremental addition of (c) spatial $\mathcal{L}_{spatial}$  and (d) ranking ($\mathcal{L}_{rank}$) loss terms to the (b) MSE ($\mathcal{L}_{MSE}$) loss term. 
  Incorporating the spatial loss term improves the spatial distribution of objects in both density maps. The density maps are further improved by the incorporation of the ranking term, which penalizes the under-estimation of counts beyond the lower-count range (top row) in the loss function. Moreover, it also helps to reduce the false positives within the lower-count range (bottom row). 
 Table \ref{tab:loss_analysis} shows the systematic improvement, in terms of mRMSE and mRMSE-nz, when integrating different terms in our loss function. The best results are obtained when integrating all three terms (classification, spatial and count) in our loss function. We also evaluate the influence of $\lambda$,  which controls the relative weight of the ranking loss. We observe $\lambda=0.1$ provides the best results and fix it for all datasets. 
\HMCrev{As mentioned in Sec.~\ref{sec:categoryspecificdensityMap}, we integrate different loss terms using  
a two-stage training strategy, which also helps to  
achieve fair  performance for the object categories that were not available during ImageNet pre-training of the backbone. 
For example,  LC framework achieves an average mRMSE-nz error of  1.88 across the COCO object categories that are unavailable in the  ImageNet dataset.}  
\begin{table}[t]
\centering
\resizebox{\columnwidth}{!}{
\begin{tabular}{@{}c|ccc@{}}
\hline
 & ${\mathcal{L}}_{class}$+${\mathcal{L}}_{MSE}$ &${\mathcal{L}}_{class}$+${\cal{L}}_{spatial}$+${\mathcal{L}}_{MSE}$ & $\mathcal{L}$ ($\lambda=0.1$) \\ \hline
mRMSE &  0.36& 0.33 & \best{0.29} \\ \hline
mRMSE-nz & 1.52 & 1.32 & \best{1.14} \\ \hline
\end{tabular}
}
\end{table}
\begin{table}[t]
\centering
\resizebox{\columnwidth}{!}{
\begin{tabular}{@{}c|ccccc@{}}
\hline
 & $\mathcal{L}$ ($\lambda=0.01$) &$\mathcal{L}$ ($\lambda=0.05$) &$\mathcal{L}$ ($\lambda=0.1$)  &$\mathcal{L}$ ($\lambda=0.5$) &  $\mathcal{L}$ ($\lambda=1$) \\ \hline
mRMSE &  0.31& \sbest{0.30} & \best{0.29} & {0.32} &0.36\\ \hline
mRMSE-nz & 1.27 & \sbest{1.16} & \best{1.14}& 1.23&1.40 \\ \hline
\end{tabular}
}\vspace{0.4em}
\caption{\HMC{Top: Progressive integration of different terms in the loss function and their impact on the final counting performance of our LC framework on the PASCAL VOC count-test set. Bottom: Influence of the weight ($\lambda$)  of ranking loss.}}
\label{tab:loss_analysis}
\end{table}

\begin{figure}[t]
\includegraphics[width=0.9\linewidth, keepaspectratio,clip=true, trim=0.2cm 10.8cm 14.9cm 0cm]{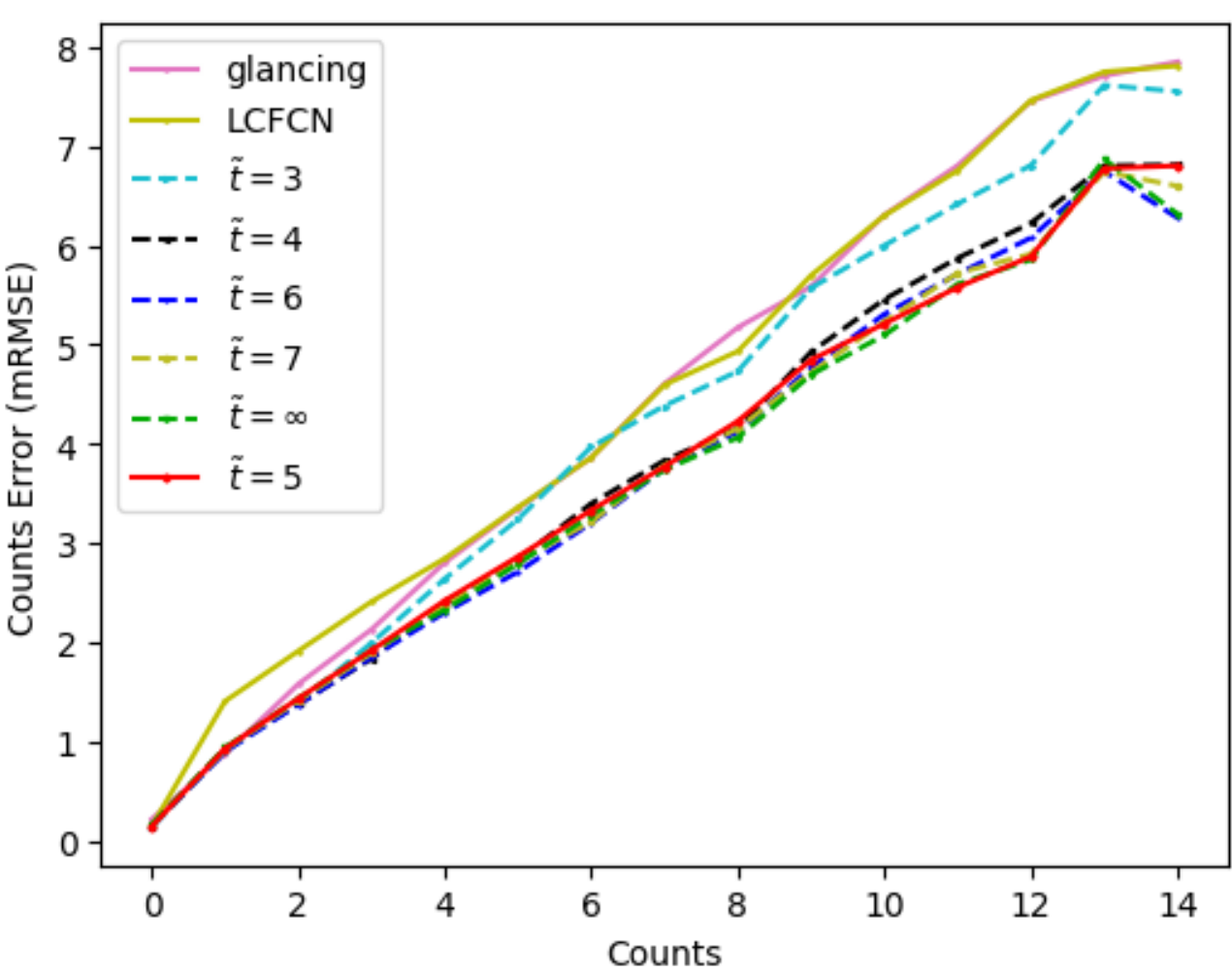}\vspace{-0.25cm}
\caption{Impact of  lower-count range on the counting performance, evaluated on the COCO count-test set. The lower-count range is defined from the count 1 till  $\tilde{t}-1$.  We vary the value of $\tilde{t}$ and plot mRMSE w.r.t. the ground-truth counts. As can be observed,  very small value such as  $\tilde{t}=3$ (cyan) leads to considerable reduction in accuracy, compared to  $\tilde{t}=5$ (red). At larger counts between 8 to 12 (x-axis),  $\tilde{t}=4$ (black) leads to a slight reduction in performance compared to $\tilde{t}=5$.  Nearly the same performance is obtained for $\tilde{t}=5$,  $\tilde{t}=6$ and $\tilde{t}=7$, indicating that the optimum balance between the annotation cost and  counting performance is  achieved at $\tilde{t}=5$. We also trained the proposed model using image-level count supervision (IC), by removing the ranking loss term from the proposed loss function, shown as $\tilde{t}=\infty$~(green). Although  $\tilde{t}=\infty$ gives the best performance, it requires higher supervision (more costly) to annotate the counts. \HMC{The figure also shows that at different ground-truth counts, the proposed method outperforms  methods using IC (glancing \cite{Chattopadhyay_2017_CVPR}) and PL (LCFCN \cite{WhereAreBlobsECCV18}) supervision.}}
\label{fig:ablation_subitizingRange}
\end{figure}

\noindent\textbf{Effect of Lower-count Range: }
In our LC and RLC frameworks, object count information beyond the lower-count range is not used for all object categories, i.e, for object categories  with ground truth counts ($t_c \ge \tilde{t}$). Psychological studies show that humans require less time to count a smaller number of instances. This property can be used to largely reduce the count annotation cost by instructing the annotators not to count beyond the lower-count range, \ie not to count more than  $\tilde{t}$ instances of the same object category. 
In Fig.~\ref{fig:ablation_subitizingRange}, we  analyze the influence of  $\tilde{t}$ on the   counting performance of our LC framework, on COCO count-test set, by varying  $\tilde{t}$ from $3$ to $7$ and plotting the count error (RMSE) at various ground-truth counts. 
The figure shows that   $\tilde{t}=5$, $\tilde{t}=6$, and $\tilde{t}=7$ give nearly the same counting performance, and hence an optimum balance between the annotation cost and performance can be obtained at~$\tilde{t}=5$. 

\noindent\textbf{Evaluation of Density Map: }
\begin{table}[t]
\centering
\resizebox{1\columnwidth}{!}{
\begin{tabular}{>{\centering\arraybackslash}p{3cm}|c|cccc}
\hline 
 Approach &SV& MAE/GAME (0)  & GAME (1) & GAME (2) & GAME (3)  \\\hline
CSRnet~\cite{CSRnetDialatedConv_2018_CVPR} & PL & 1.2 & 1.7 & 2.2 & 2.44 \\ \hline
LCFCN \cite{WhereAreBlobsECCV18}& PL & 0.72 & 1.41 & 2.12 & 2.8 \\ \hline
Proposed & LC & 0.71 & 1.14 & 1.5 & 1.83\\ \hline
\end{tabular}}
\vspace{0.02em}
\caption{{Density map evaluation on  person category  of PASCAL VOC 2007 count-test set, in terms of  grid average mean absolute error (GAME)  and mean absolute error (MAE or GAME (0)) metrics. Improved GAME (3) performance indicates that our density map accurately predicts the spatial distribution of objects, compared to the crowd counting-based method \cite{CSRnetDialatedConv_2018_CVPR} and localization-based method \cite{WhereAreBlobsECCV18}.}}
\label{tab:density_pascal}
\vspace{-0.2cm}
\end{table}
We employ the standard grid average mean absolute error (GAME) evaluation metric \cite{GAME} used in crowd counting to evaluate spatial distribution consistency in the density map. In GAME (n), an image is divided into $4^n$ non-overlapping grid cells. The mean absolute error (MAE) between the predicted  and the  ground-truth local counts are reported for  $n=0,~1,~2$ and $3$, as in \cite{GAME}.  We compare our approach with the state-of-the-art PL supervised counting approach (LCFCN) \cite{WhereAreBlobsECCV18} on the 20 categories of the PASCAL VOC 2007 count-test set. Furthermore, we also compare with a popular crowd counting approach (CSRnet) \cite{CSRnetDialatedConv_2018_CVPR} on the person category of the PASCAL VOC 2007 by retraining it on this dataset.  For the person category, the PL supervised LCFCN and CSRnet approaches achieve scores of $2.80$ and $2.44$ in GAME (3) \HMC{as shown in Table~\ref{tab:density_pascal}.}  
The proposed method outperforms LCFCN and CSRnet in GAME (3) with a score of $1.83$, demonstrating the capability of our approach in estimating the precise spatial distribution of object counts. Moreover, our method outperforms LCFCN for all 20 categories.  Additional ablation  results are available in the supplementary material.\\ 
\HMC{\noindent\textbf{Evaluation of  Hard Spatial Guidance:}}
\HMC{We further validate the effectiveness of the proposed hard spatial guidance strategy by using the ground-truth point-level annotations \cite{WhereAreBlobsECCV18} as the spatial mask  $\textbf{B}^{c}$ for training the density branch. Tab.~\ref{tab:LC_withPL} shows that  only a slight performance gain can be achieved using  this higher supervision level, which indicates the effectiveness of the proposed hard spatial guidance strategy. Further, note that point-level annotations (\eg the center of the bounding-box used in LCFCN \cite{WhereAreBlobsECCV18})  are semantically less consistent for learning the density map in a natural scene having large intra-class scale and pose variations. On the other side, the proposed spatial mask  $\textbf{B}^{c}$ often points to the semantically meaningful discriminative object regions (peaks) already captured by the image classifier branch, leading to an  accuracy  comparable to using ground-truth point-level supervision in our LC framework.  
This experiment also validates the advantage of the proposed architecture compared to  LCFCN \cite{WhereAreBlobsECCV18} using identical point-level supervision and ResNet50 backbone. It is worth mentioning that the proposed framework operates at a three  times faster inference speed compared to  LCFCN \cite{WhereAreBlobsECCV18}. } 
\begin{table}[t]
\resizebox{\columnwidth}{!}{
\begin{tabular}{>{\centering\arraybackslash}p{3cm}|c|cccc}
\hline 
 Approach &SV&mRMSE&  \begin{tabular}[c]{@{}c@{}}mRMSE\\-nz\end{tabular}   & \begin{tabular}[c]{@{}c@{}} m-rel\\RMSE\end{tabular}   &\begin{tabular}[c]{@{}c@{}} m-rel\\RMSE-nz \end{tabular} \\\hline
Proposed (LC+point) & PL & 0.28 & 1.12 & 0.16 & 0.58 \\\hline
Proposed (LC) & LC & 0.29 & 1.14 & 0.17 & 0.61 \\\hline
\end{tabular}
}\vspace{0.4em}
\caption{\HMC{Evaluation of the  hard spatial guidance  on the PASCAL VOC count-test. Only a slight performance gain is obtained by integrating the point-level supervision for the spatial guidance in the proposed LC framework, indicating the localization capability of our  hard spatial guidance strategy.}} 
\label{tab:LC_withPL}
\vspace{-0.2cm}
\end{table}

 \subsubsection{Counting with Reduced Lower-count Supervision (RLC)}
\noindent \HMC{\textbf{Influence of $|\mathcal{A}|$:}} To evaluate the significance of the additional components introduced in our RLC framework, we report an ablation study on the COCO \cite{coco_eccv2014} dataset. First, we show how the number of count-annotated categories ($|\mathcal{A}|$) affects the overall category-specific accuracy. We start with only using count annotations of 20 categories, and then gradually increase the number of count-annotated categories to 40, 60 and 80. Note that the compared methods from our LC  and LCFCN \cite{WhereAreBlobsECCV18} use counts for all 80 categories. The comparison is shown in Table~\ref{tab: accu_number_A}. With only  half the number of count-annotated categories (40/40 split), our RLC approach performs better than LCFCN \cite{WhereAreBlobsECCV18}, which requires point-level supervision.
We can see that our method obtains consistent performance gain with the increase of supervision. Furthermore, when using count annotations  for 60 categories, we obtain an mRMSE of 0.36, which is comparable with the 0.34 of our LC framework, which uses lower-count supervision for all categories. 
\begin{SCtable}[][t]
\centering
\resizebox{0.53\columnwidth}{!}{
\begin{tabular}{c|c|c}
\hline
Method & Split & mRMSE  \\ \hline
\multirow{4}{*}{Proposed (RLC)} & 20/60 & 0.41  \\ \cline{2-3}
  & 40/40 & 0.37 \\ \cline{2-3}
  & 60/20 & {0.36} \\ \cline{2-3}
  & 80/0 & \sbest{0.35} \\ \cline{2-3}\hline
  LCFCN  \cite{WhereAreBlobsECCV18} & 80/0 & 0.39 \\ \hline
Proposed (LC)  & 80/0 & \best{0.34}  \\ \hline
\end{tabular}}\vspace{-0.5em} 
\caption{Evaluation of RLC framework with different known/unknown count splits, on the COCO count-test set. \HMC{At 60/20 split  RLC model results are comparable to LC model.}}
\label{tab: accu_number_A}
\vspace{-0.2cm}
\end{SCtable}
 
\begin{table}[t]
\resizebox{1.0\columnwidth}{!}{
\begin{tabular}{c|c|c|c c}
\hline
Design  & \begin{tabular}[c]{@{}c@{}}Size of \\ hidden layer\end{tabular}& Activation  & mRMSE         & mRMSE-nz     \\ \hline
2 layers & $P/2$         & Relu      & {0.36} & 1.97          \\ \hline
2 layers & $P/2$         & Leaky Relu      & {0.36} & \sbest{1.96}          \\ \hline
2 layers &  $P/2$  & Softmax    & {0.36} & \best{1.94} \\ \hline
2 layers & $P$         & Softmax  & 0.37          & 1.98          \\ \hline
3 layers & $P$         & Softmax     & 0.38          & 2.07          \\ \hline
\end{tabular}}\vspace{0.2em}
\caption{Study on the structure of the weight modulation layer on the COCO count-test set by  varying  the number of fully connected layers,  size of hidden layers,  and  activation  functions. We observe that the  best  performance (mRMSE-nz) is obtained by a  2-layered network  with a bottleneck low-dimensional embedding (to $\frac{P}{2}$) and  a softmax  non-linear activation,  and fix it for all datasets.}
\label{tab:ablation_transfer}
\vspace{-0.3cm}
\end{table}

\noindent\HMC{\textbf{Optimal Architecture for Weight Modulation:}}
 In Table~\ref{tab:ablation_transfer}, we perform an experiment to identify the optimum structure for the weight modulation layer, by varying the number of fully connected layers, size of the hidden layer and activation functions. 
We find that a 2-layer network with a bottleneck low-dimensional embedding (to $\frac{P}{2}$), and using softmax activation works best and therefore follow this structure in all our experiments.

\begin{table}[t]
\resizebox{\columnwidth}{!}{
\begin{tabular}{>{\centering\arraybackslash}p{3cm}|c|cccc}
\hline 
 Approach &SV&mRMSE&  \begin{tabular}[c]{@{}c@{}}mRMSE\\-nz\end{tabular}   & \begin{tabular}[c]{@{}c@{}} m-rel\\RMSE\end{tabular}   &\begin{tabular}[c]{@{}c@{}} m-rel\\RMSE-nz \end{tabular} \\\hline
Aso-sub-ft-3$\times$3 \cite{Chattopadhyay_2017_CVPR} & BB & 0.38 & 2.08 & 0.24 & 0.87 \\\hline
Seq-sub-ft-3$\times$3 \cite{Chattopadhyay_2017_CVPR} & BB & \sbest{0.35} & \sbest{1.96} & \best{0.18} & \sbest{0.82} \\\hline
ens \cite{Chattopadhyay_2017_CVPR} & BB & 0.36 & 1.98 & \best{0.18} & \best{0.81} \\\hline
Fast-RCNN \cite{Chattopadhyay_2017_CVPR} & BB & 0.49 & 2.78 & 0.20 & 1.13 \\\hline
LCFCN \cite{WhereAreBlobsECCV18} & PL & 0.38 & 2.20 & {0.19} & 0.99 \\\hline
glancing \cite{Chattopadhyay_2017_CVPR} & IC & 0.42 & 2.25 & 0.23 & 0.91 \\\hline
Proposed (LC) & LC & \best{0.34} & \best{1.89} & \best{0.18} & 0.84\\\hline
Proposed (RLC)        & RLC &    0.36$\pm$0.01 &       \sbest{1.96$\pm$0.01}     &  {0.19$\pm$0.00}   &  {0.84$\pm$0.01}     \\ \hline\hline
Proposed (RLC)        & LC &    0.35 &       {1.90}     &  {0.18}   &  {0.83}     \\ \hline
\end{tabular}
}\vspace{0.2em}
\caption{State-of-the-art counting performance comparison on the COCO count-test set. 
Despite using reduced supervision, our LC approach provides superior results compared to existing methods on three metrics. Compared to the image-level count (IC) supervised approach \cite{Chattopadhyay_2017_CVPR}, our LC method achieves an absolute gain of 8\% in terms of mRMSE.   For the RLC framework, we follow a 60/20 known/unknown count annotation split, 
and repeat experiments three times, randomly interchanging object categories among these sets,  and report the mean and standard deviation.  Note that  despite using only the category labels for 20 categories, \HMC{our RLC framework performs favorably compared to its upper-bound (last row) as well as  its LC counterpart}. } 
\label{tab:counting_coco}
\vspace{-0.15cm}
\end{table}

   \begin{figure}[t]
		\centering	\includegraphics[width=0.9\linewidth, clip=true, trim=1.0cm 0.1cm 0.1cm 0.2cm]{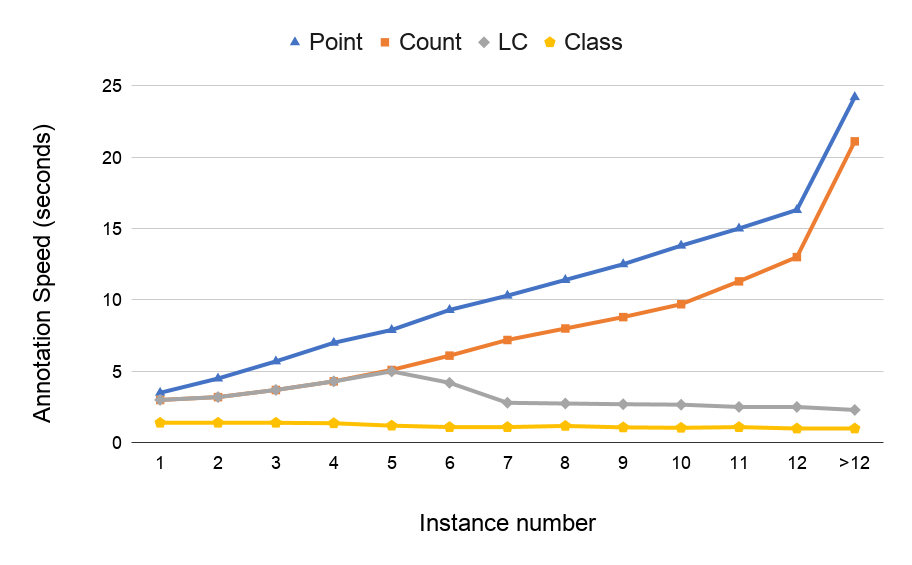}\\  \vspace{-0.35cm}
			\caption{\HMC{Annotation speed vs number of per-category instances in an image. The  cost of the proposed LC annotation is  favorable at larger count ranges, compared to  point and count annotations.}}
			\label{Graph_InstanceCount_vs_speed}
\end{figure}

\HMC{As mentioned in Sec.~\ref{subsec:weightModulationLayer}, backpropagation is not performed  from $\Psi$ to the image classifier (shown with green line in Fig.~\ref{fig:SPC_archit}). This is because the  count annotations required to train $\Psi$ are available only for the set $\mathcal{A}$, which can limit the generalizability of $\Psi$ to perform counting on set $\mathcal{B}$.
Further, training the $\bm{w}^c_{cls}$ with additional counting objective may lead to inferior image classification accuracy on the set $\mathcal{A}$. To avoid these issues, we do not perform the backpropagation from $\Psi$ to the image classifier branch. We further validate this hypothesis by introducing the backpropagation to the image classification branch, and observe that the performance on COCO reduced to an mRMSE error of 0.38 on a $60/20$ split, instead of its original mRMSE 0.36 in Tab.~\ref{tab: accu_number_A}.} 

 \noindent\HMC{\textbf{Influence of $\mathbf{D}^{tot}$:} To further investigate the influence of  additionally introduced total density map $\mathbf{D}^{tot}$ in our RLC framework,  we retrain the framework on COCO dataset by  removing the multiplication with the total density map $\mathbf{D}^{tot}$ during the computation of category-specific density map. We observe a performance reduction in this new setting where [mRMSE , mRMSE-nz, m-relRMSE, m-relRMSE] errors are increased to  [0.37, 1.99, 0.20, 0.86] instead of its original values [0.36, 1.96, 0.19, 0.84] ({see Tab.~\ref{tab:counting_coco}}), demonstrating the importance of total density map $\mathbf{D}^{tot}$ in our RLC setting.}

\noindent\HMC{\textbf{Comparison of annotation speeds:}
To further validate the annotation speed-up obtained through the proposed LC and RLC annotation strategies, we perform an annotation speed experiment on \HMCrev{a subset of} COCO minval5k dataset. In this experiment, we evaluate the annotation cost of  different annotation strategies, on  images containing different number of instances. Fig.~\ref{Graph_InstanceCount_vs_speed} shows the  average time required for annotating an object category \HMCrev{(person category)} in an image for point per instance, image-level count (IC), LC, and image-level class annotations.  Point-level annotation of each additional instance in an image takes around 1.1 seconds. The count annotations are faster at smaller count range  (costs nearly 0.5 seconds to count an additional instance, until count 4) and slower at larger count ranges. At larger count ranges,  the annotator has to fixate each instance for counting, which costs nearly 1 second to count an additional instance.  For both point and count annotations, the annotation time is proportional to the number of instances. }

\HMC{Our LC annotation cost  follows the count annotation cost at the initial count ranges (until instance count 5), but interestingly, the annotation cost is observed to reduce at higher counts.   
This reduced annotation cost is intuitive,  since the annotator can quickly identify $>$4 counts, with a single glance at the image, without actually counting the number of instances. A similar speed-up is observed for the category-level annotations at larger counts (count-1 at 1.4 seconds vs count 12 at 1.1 seconds), since it is quicker for the annotator to spot atleast one instance of the object category on such images. }

\HMC{The RLC annotation cost can be derived from the LC annotation and image-level class annotation costs, since RLC annotation requires LC annotation for some categories and image-level class annotation for the remaining categories. Depending on the number of count-labeled categories, the RLC annotation cost varies between the class and LC annotation costs. In summary, different to the  point-level and image-level count annotations whose annotation cost increases proportional to the number of instances, LC and RLC annotation costs do not increase beyond the instance count 5.  Further, for all types of  annotation (class, count, point-level, LC and RLC), the annotation cost is  proportional to the number of categories.}

\subsection{State-of-the-art Comparison}
Here, we compare the counting performance of our LC and RLC frameworks with state-of-the-art approaches.  At first, we evaluate category-specific object counting on COCO, Visual Genome and  PASCAL VOC 2007  datasets and the results are shown in Tables  \ref{tab:counting_coco}, \ref{tab:results_vg1} and \ref{tab:counting_pascal}, respectively. Secondly, we evaluate category-independent total  object counting performance on COCO, Visual Genome datasets and the results are compared with existing methods in Tables \ref{tab:results_coco_total_count}, \ref{tab:results_vg_total}, respectively.   Finally, we evaluate the generalization ability of category-independent counts  predicted by our RLC framework, to  unsupervised object categories (\ie,~without category and count annotations), and the results are compared with  methods using count-supervision in Table \ref{tab:results_coco_10}. Additional  results and failure cases are included in the supplementary material.

\subsubsection{Category-specific counting}
Table \ref{tab:counting_coco} compares the  category-specific object counting  performances on the COCO dataset, which  has 80 object categories. To train our RLC framework, we use  a 60/20 split, where the 80 categories are randomly split into 60 and 20 categories.  Here, 60 categories have both class labels and count annotations, and 20 categories only have class labels. 
To avoid any bias due to a particular random split, we repeat experiments three times with different random splits and report the mean and standard deviation in the table. \HMC{In addition, we train our RLC framework using LC  supervision (\ie, 80/0 split) for obtaining the upper-bounds of the architecture.}   
For all other methods, both count (lower-count for LC) information and class labels are used for all 80 object categories. 
Among all existing methods, the two BB supervised approaches (Seq-sub-ft-3x3 and ens) yield mRMSE scores of $0.35$ and $0.36$, respectively. The PL supervised LCFCN approach \cite{WhereAreBlobsECCV18} achieves an mRMSE score of $0.38$. The IC supervised glancing approach obtains an  mRMSE score of $0.42$. Our LC framework outperforms the glancing approach with an absolute gain of 8\% in mRMSE. Furthermore, our LC framework  also provides consistent improvements over the glancing approach, in the other three error metrics and is only below the two BB supervised methods (Seq-sub-ft3x3 and ens) in m-relRMSE-nz. \HMC{Further, our RLC approach, despite using only  category-level binary labels for 20 categories, performs favorably compared to its upper bound (last row) as well as its LC counterpart.}

\begin{figure}[t]
		\centering
        \includegraphics[width=1\columnwidth, clip=true, trim=0cm 4.3cm 27cm 0cm]{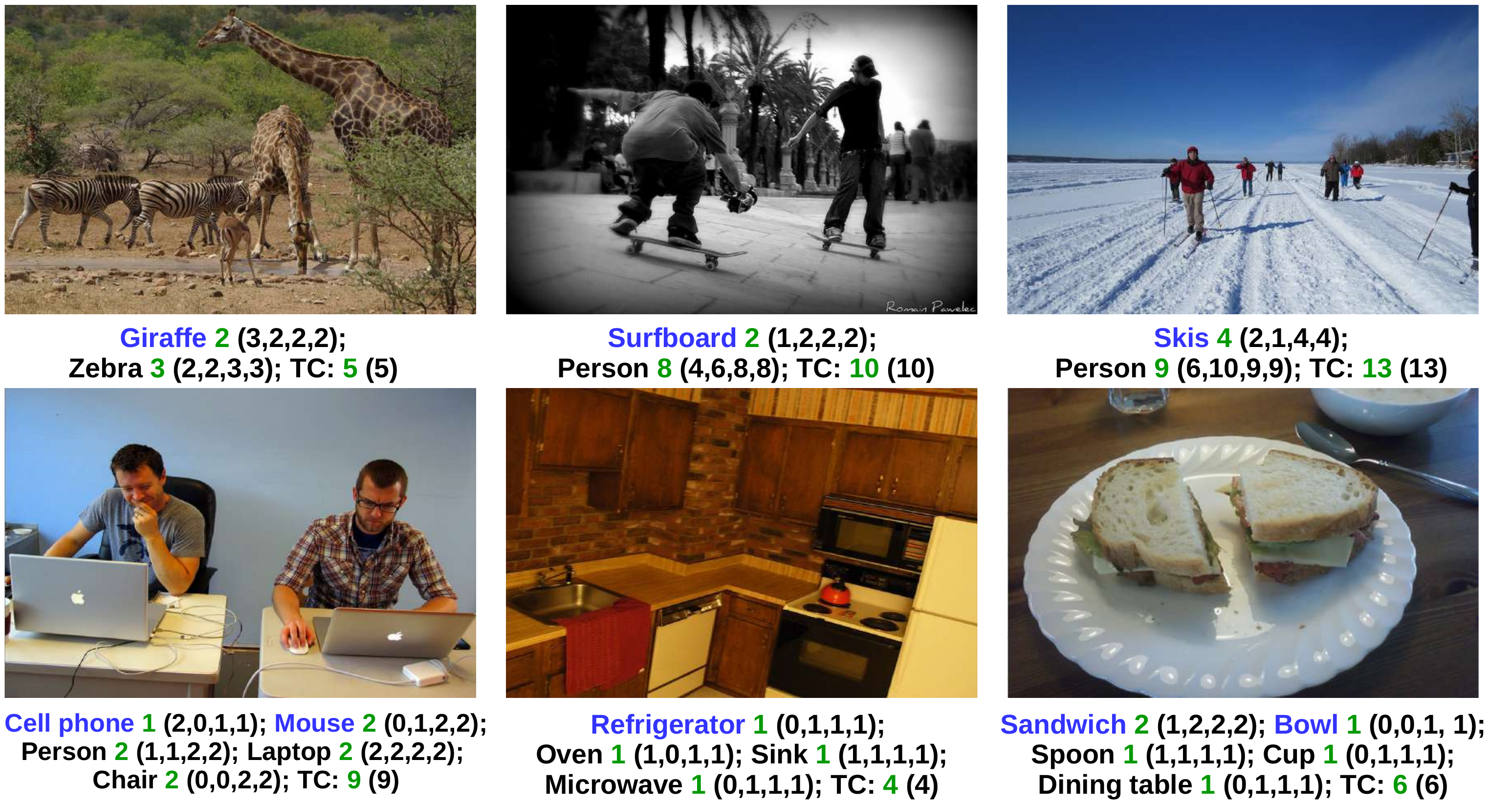}\vspace{-0.2cm}
         \caption{Object counting examples on the COCO and Visual Genome datasets. The ground-truth is shown in \textcolor{green}{green}, while the predictions  by glancing \cite{Chattopadhyay_2017_CVPR}, LCFCN \cite{WhereAreBlobsECCV18},  our LC framework, and our RLC  framework, are shown sequentially  inside the parentheses.  The examples show that our LC and RLC frameworks accurately  predict counts   for diverse categories (animals  to food items), and even beyond the lower-count range. Although the count-annotation of object categories indicated with \textcolor{blue}{blue} are not used for   training the RLC framework, their counts are predicted accurately. Finally,  the category-independent total count (TC) predicted by our RLC framework is shown separately, inside parentheses. Best viewed in zoom. }
		\label{Fig:qual_coco}
\end{figure}

For the large-scale Visual Genome dataset, our proposed RLC method is evaluated under two different splits. The first split follows \cite{zero_shot_eccv18}: all 609 classes are split into 479 and 130 classes, where 479 classes have both image labels and counts annotations, and 130 classes only have image labels. For the second split, we experiment on a more challenging case: the 609 classes are randomly split into half and half (304 classes and 305 classes), where the first half has both image labels and count annotations, and the second half only has image labels. The comparisons are shown in Table \ref{tab:results_vg1}. It shows that both our proposed LC and RLC methods perform on par with existing methods (glancing and LCFCN). Note that in the case of our RLC approach, count information for some classes is not used during training. Fig. \ref{Fig:qual_coco} shows object counting examples on the COCO and Visual Genome datasets, using glancing \cite{Chattopadhyay_2017_CVPR}, LCFCN \cite{WhereAreBlobsECCV18} and the proposed LC,  RLC frameworks. Both our approaches perform accurate counting on various categories (animals to food items) under challenging situations.

On the PASCAL VOC 2007 dataset, we only evaluate our LC method since RLC is designed for large-scale datasets and PASCAL VOC only has 20 classes. Table \ref{tab:counting_pascal} shows that the glancing approach  of \cite{Chattopadhyay_2017_CVPR} using image-level supervision both within and beyond the lower-count range (IC) achieves an mRMSE score of $0.50$. Our LC supervised approach outperforms the glancing method with an absolute gain of 21\% in mRMSE. Furthermore, our approach achieves favorable performance  on all error metrics, compared to the state-of-the-art point-level and bounding box based supervised methods.

\begin{table}[t]
\centering
\resizebox{1\columnwidth}{!}{
\begin{tabular}{c|c|c|c|c|c}
\hline
\multirow{2}{*}{Metrics} & \multirow{2}{*}{Glancing \cite{Chattopadhyay_2017_CVPR}}& \multirow{2}{*}{LCFCN \cite{WhereAreBlobsECCV18}} & \multirow{2}{*}{Proposed (LC)} & \multicolumn{2}{c}{Proposed (RLC)} \\ \cline{5-6}
 &           &      &     & split 304/305  & split 479/130  \\ \hline
 SV                        &  IC &  PL   & LC  & RLC          & RLC        \\ \hline 
mRMSE                        &  \best{0.15} &  0.16   & \best{0.15}  & \best{0.15}           & \best{0.15}           \\ \hline 
mRMSE-nz       &  1.57  &   1.62  & \best{1.51}   & 1.54           & \sbest{1.53}           \\ \hline
m-relRMSE           &  \best{0.09}   &  \best{0.09}    & \best{0.09}  & 0.10           & \best{0.09}           \\ \hline
\end{tabular}}\vspace{0.2em}
\caption{Results on  Visual Genome dataset for  all classes. Our approaches perform favorably against methods using higher levels of supervision.  Our  RLC framework is evaluated on two different known/unknown count splits and, for both splits,  performs   on par with its LC counterpart which  uses lower-count annotations for all object categories. }
\label{tab:results_vg1}
\vspace{-0.35cm}
\end{table}

\begin{table}[t]
\resizebox{\columnwidth}{!}{
\begin{tabular}{>{\centering\arraybackslash}p{2.8cm}|c|cccc}
\hline 

 Approach &SV&mRMSE&  \begin{tabular}[c]{@{}c@{}}mRMSE\\-nz\end{tabular}   & \begin{tabular}[c]{@{}c@{}} m-rel\\RMSE\end{tabular}   &\begin{tabular}[c]{@{}c@{}} m-rel\\RMSE-nz \end{tabular}
    \\\hline
Aso-sub-ft-3$\times$3 \cite{Chattopadhyay_2017_CVPR} & BB & 0.43 & 1.65 & 0.22 & 0.68 \\\hline
Seq-sub-ft-3$\times$3 \cite{Chattopadhyay_2017_CVPR} & BB & 0.42 & 1.65 & 0.21 & 0.68 \\\hline
ens \cite{Chattopadhyay_2017_CVPR} & BB & 0.42 & 1.68 & 0.20 & 0.65 \\\hline
Fast-RCNN \cite{Chattopadhyay_2017_CVPR} & BB & 0.50 & 1.92 & 0.26 & 0.85 \\\hline
LCFCN \cite{WhereAreBlobsECCV18} & PL & \sbest{0.31} & \sbest{1.20} & \best{0.17} & \best{0.61} \\\hline
LC-PSPNet \cite{WhereAreBlobsECCV18} & PL & 0.35 & 1.32 & 0.20 & 0.70 \\\hline
glancing \cite{Chattopadhyay_2017_CVPR} & IC & 0.50 & 1.83 & 0.27 & 0.73 \\\hline
Divide-count \cite{GenericDivideCount}& IC & 0.51 & - & - & -  \\\hline
Proposed (LC)  & LC & \best{0.29} & \best{1.14} & \best{0.17} & \best{0.61}\\\hline
\end{tabular}
}\vspace{0.2em}
\caption{State-of-the-art counting performance comparison on the Pascal VOC 2007 count-test. Our LC supervised approach outperforms existing methods.}
\label{tab:counting_pascal}
\vspace{-0.2cm}
\end{table}

\subsubsection{Category-independent Total Object Counting}
For the COCO dataset, we report the performance of category-independent total count prediction in Table \ref{tab:results_coco_total_count}. Our RLC method has a dedicated category-independent density estimation sub-branch that can directly provide the total count predictions. For methods that do not have such a capability, we compute the total count by summing up the category-specific counts. 
We note that, in the presence of highly complex scenes with a diverse range of object categories, difficult object instances are missed during category-specific counting. When such errors occur for several classes, they get accumulated in the total object count. As a result, the total counts have a larger variation from the ground-truth. In such cases, a holistic density map helps in estimating the overall object count. This is evident from Table \ref{tab:results_coco_total_count}, where the proposed  category-independent density estimation sub-branch of the RLC framework predicts the most accurate total counts on the COCO dataset \HMC{(while trained with RLC or LC data)}.

\begin{table}[t]
\resizebox{1\columnwidth}{!}{
\begin{tabular}{c|c|c|c|c||c}
\hline
        &glancing\cite{Chattopadhyay_2017_CVPR}  & \begin{tabular}[c]{@{}c@{}}LCFCN \cite{WhereAreBlobsECCV18} \end{tabular} & \begin{tabular}[c]{@{}c@{}}Proposed (LC)  \end{tabular} & 
         \begin{tabular}[c]{@{}c@{}}Proposed (RLC) \end{tabular}&
        \begin{tabular}[c]{@{}c@{}}Proposed (RLC) \end{tabular} \\ \hline
        SV & IC &  PL& LC & RLC & LC \\ \hline    
Categ.-indep. & \xmark &  \xmark & \xmark  &\cmark &\cmark \\ \hline    
mRMSE & 7.01$\pm$ 0.00  &  6.13$\pm$ 0.00 &  \sbest{4.4$\pm$0.00}  &  \best{4.28$\pm$0.03} &  4.00$\pm$0.00\\ \hline
m-relRMSE  &  2.05$\pm$ 0.00 &   1.65$\pm$ 0.00  & \sbest{1.14$\pm$0.00}  &   \best{1.06$\pm$ 0.01} &1.01$\pm$0.00 \\ \hline
\end{tabular}}\vspace{0.2em}
\caption{Results of  total count estimation on  COCO dataset. Note that our RLC framework does not use the count annotations of 20 classes during training and performs favorably \HMC{compared to  its upper bound (last column) as well as other methods using count annotations for all classes, demonstrating its generalizability.}}
 \label{tab:results_coco_total_count}
\end{table}

For the Visual Genome (VG) dataset, the results for total counting are shown in Table~\ref{tab:results_vg_total}. The VG dataset has high diversity in terms of object classes, which makes it a very challenging dataset. Furthermore, each image has a large number of objects, i.e., 35 object categories/image on average, which makes the counting task even more difficult. Remarkably, for both splits, our proposed category-independent density sub-branch achieves the best  performance despite the fact  that the counts for the 305 or 130 classes (depending on split) are not used during training. The favorable performance of our RLC framework compared to its LC counterpart demonstrates the  generalization ability of the RLC framework on large-scale datasets.
\begin{table}[t]
\centering
\resizebox{1\columnwidth}{!}{
\begin{tabular}{c|c|c|c|c|c}
\hline
\multirow{2}{*}{Method} & \multirow{2}{*}{glancing \cite{Chattopadhyay_2017_CVPR}} & \multirow{2}{*}{LCFCN \cite{WhereAreBlobsECCV18}} & \multirow{2}{*}{Proposed (LC)} & \multicolumn{2}{c}{Proposed (RLC)}                                         \\ \cline{5-6} 
                  &            &&          & \multicolumn{1}{c|}{split 304/305} & \multicolumn{1}{c}{split 479/130} \\ \hline
Categ.-indep. & \xmark  & \xmark &  \xmark &  \cmark     &       \cmark           \\ \hline
mRMSE            & 9.88 &13.20 & 9.49  & \sbest{7.35}  & \best{6.93} \\ \hline
m-relRMSE       & 2.40 & 3.12 & 2.34           & \sbest{1.97}            & \best{1.94}            \\ \hline
\end{tabular}}\vspace{0.2em}
\caption{Results of total count estimation on the Visual Genome dataset. On this large-scale dataset, our LC and RLC frameworks demonstrate superior performance compared to existing approaches.  Our RLC framework is evaluated on two challenging known/unknown count splits, and for both splits, the total counts predicted by our RLC framework, which has a dedicated category-independent density sub-branch, are more accurate, compared to the total count estimated using the LC framework.}  
\label{tab:results_vg_total}
\vspace{-0.35cm}
\end{table}

\subsubsection{Generalization to Unsupervised Object Counting} 
\vspace{-0.05cm}
Here, we consider a  learning scenario that is motivated by the zero-shot  setting  \cite{zero_shot_eccv18}, where  80 COCO classes are split  into subsets of 60-10-10. The first 60 classes have both image-level labels and object count annotations. The second 10 classes only have image-level labels but no count annotations. The last 10 classes do not have any annotations and images containing objects from any of those 10 categories  are not used during  the end-to-end training of our RLC framework. 
We train our RLC method using the first two sets (60+10 classes) with labels, and we test the category-independent density branch on images that only have objects from the last 10 classes. The total count performance on the last 10 classes is shown in Table \ref{tab:results_coco_10}. Note that all other methods, including our LC framework, are trained  using count/lower-count annotations for all 80 object categories. 
We can see that, for the  unannotated categories, our RLC framework performs favorably compared  to LCFCN \cite{WhereAreBlobsECCV18}, 
which requires point-level annotations for training.
Our RLC method achieves 1.42 and 0.85, in terms of mRMSE and m-relRMSE, while LCFCN  obtains 1.59 and 0.87. Our RLC setting  even performs comparably to our LC  setting in mRMSE. Since the last 10 categories are not  used during our training, the favorable performance demonstrates that the proposed RLC method can generalize to predict the total count of un annotated categories.  
\begin{table}[t]
\centering
\resizebox{1\columnwidth}{!}{
\begin{tabular}{c|c|c|c|c}
\hline
        & glancing\cite{Chattopadhyay_2017_CVPR} & \begin{tabular}[c]{@{}c@{}}LCFCN \cite{WhereAreBlobsECCV18}\end{tabular} & \begin{tabular}[c]{@{}c@{}}Proposed (LC)\end{tabular} & \begin{tabular}[c]{@{}c@{}}Proposed (RLC) \end{tabular} \\ \hline
Categ.-indep. &\xmark& \xmark & \xmark & \cmark\\ \hline
mRMSE     &2.66& 1.59                                                           & \sbest{1.48}                                                          & \best{1.42}                                                                \\ \hline
m-relRMSE &1.60& {0.87}                                                           &\best{0.76}                                                          & \sbest{0.85}                                                               \\ \hline
\end{tabular}
}\vspace{0.2em}
\caption{Total  count estimation  performance of RLC framework,  on 10 unsupervised  classes of the COCO split 60-10-10. The images containing these 10 classes are not included in our RLC framework training. Both \cite{Chattopadhyay_2017_CVPR, WhereAreBlobsECCV18} and our LC setting are trained using the complete count-train set with  count annotations for all 80 categories. Our RLC approach with less supervision performs comparably to these methods.}
\label{tab:results_coco_10}
\vspace{-0.2cm}
\end{table}

 \subsection{Application to Weakly Supervised Instance Segmentation}
 \label{sec_instance_seg}
 \vspace{-0.15cm}
The category-specific density maps generated by our LC framework 
 can also be utilized for instance segmentation. Note that the local summation of an ideal density map over a ground-truth segmentation mask is nearly one. 
 We use this property to improve a recent  image-level supervised instance segmentation method (PRM \cite{PRM}). PRM employs a scoring metric that combines instance-level cues from peak response maps $R$, class-aware information from object category maps and spatial continuity priors from off-the-shelf object proposals \cite{mcg_2017,cob_eccv2016}. The peak response maps are generated from local maxima (peaks of $\tilde{\textbf{M}}^{c}$) through a peak backpropagation process \cite{PRM}. The scoring metric is then used to rank object proposals corresponding to each peak for instance mask prediction. We improve the scoring metric by introducing an additional term $d_p$.  The term  $d_p$ penalizes an object proposal $P_r$ if the predicted count in those regions of the density map $\textbf{D}^c$ is deviated from one, as $d_p$= $|1-\texttt{sum}(\textbf{D}^c\cdot P_r)|$. Here, $|~|$ is the absolute value operator. 
 For each peak, the new scoring metric  $Score$ selects the  highest-scoring object proposal $P_r$: 
\begin{equation}
\label{eq:score}
    Score=\alpha\cdot R*P_r+R*\hat{P_r}-\beta\cdot Q*P_r-\gamma\cdot d_p.
\end{equation}
Here, the background mask $Q$ is derived from the object category map and $\hat{P_r}$ is the contour mask of the proposal $P_r$, derived using a morphological gradient \cite{PRM}. Parameters $\alpha$,  $\beta$  and $\gamma$ are empirically set as in \cite{PRM}. 

Following PRM, we evaluate the instance segmentation performance on the PASCAL VOC 2012 dataset and the results are shown in Table~\ref{tab:ins_seg_pascal}.  For  fair comparison, we utilize the same proposals (MCG) as used in \cite{PRM}. Specifically, the combinatorial grouping framework of \cite{mcg_2017}  is used in conjunction with the region hierarchies of \cite{cob_eccv2016}, which  is referred to as MCG in \cite{PRM}. Note that our approach is generic and can be used with any object proposal method.  In addition to PRM, the image-level  supervised object detection methods MELM \cite{melm_18}, CAM \cite{CAM} and SPN \cite{spn_iccv2017} used with MCG   and reported by \cite{PRM} are also included in 
Table~\ref{tab:ins_seg_pascal}. 

\begin{figure}[t]
		\centering 
			\includegraphics[width=1\linewidth, keepaspectratio,clip=true, trim=0cm 15.2cm 15.5cm 0cm]{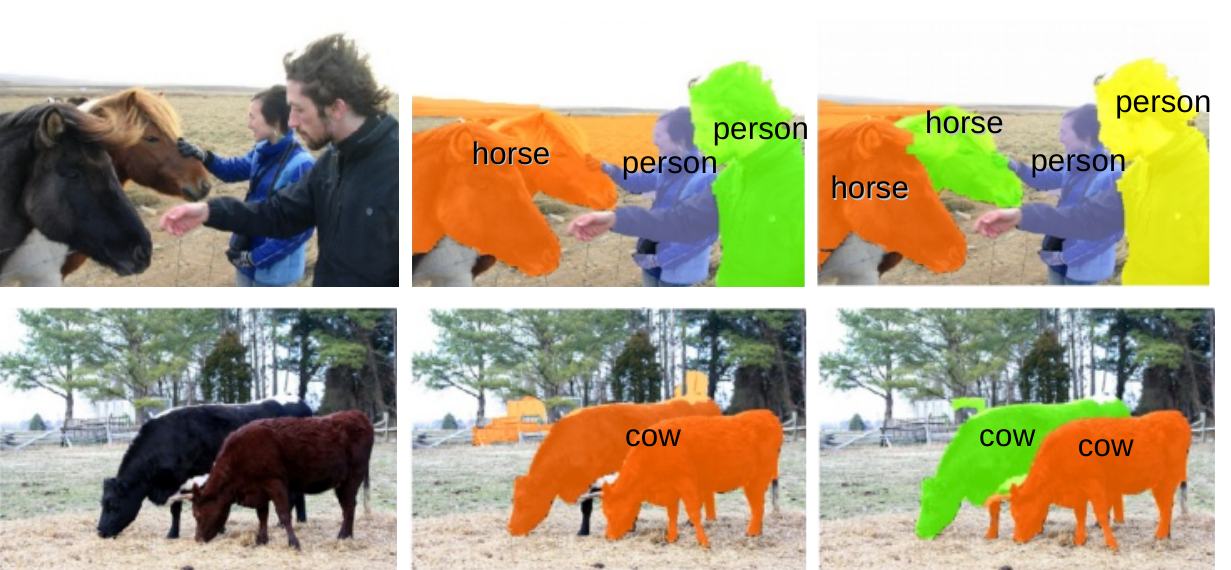}\\
						\hspace*{0.0\linewidth} (a) Input Image  \hspace*{0.07\linewidth} (b) PRM\cite{PRM} \hspace*{0.07\linewidth} (c) Our Approach  \\  \vspace{-0.2cm}
			\caption{Instance segmentation examples  of PRM \cite{PRM} and our approach. Our approach accurately delineates multiple spatially adjacent  object instances of the  horse and cow categories.}
			\label{fig:experi_instanceSeg}
\end{figure}

 The proposed method largely outperforms all the baseline approaches and \cite{PRM}, in all four evaluation metrics. Even though our approach slightly increases the supervision level (lower-count information), it improves PRM with a relative gain of 17.8$\%$ in terms of average best overlap (ABO). Compared to PRM, the gain obtained at a lower IoU threshold (0.25) highlights the improved location prediction capability of the proposed method. Furthermore, the gain obtained at a higher IoU threshold (0.75) indicates the effectiveness of the proposed scoring function in assigning a higher score to the object proposal that has the highest overlap with the ground-truth object, as indicated by the improved ABO performance. Fig.~\ref{fig:experi_instanceSeg} shows a  qualitative instance segmentation comparison between our approach and PRM. 

\begin{table}[t]
\small
 \centering
\resizebox{0.95\columnwidth}{!}{
\begin{tabular}{>{\centering\arraybackslash}p{3cm}|ccccc}
\hline
\multicolumn{2}{c|}{Method}& \multicolumn{1}{c}{$mAP^r_{0.25}$} & \multicolumn{1}{c}{$mAP^r_{0.5}$} & \multicolumn{1}{c}{$mAP^r_{0.75}$} & \multicolumn{1}{c}{ABO} \\\hline 
\multicolumn{2}{c|}{MELM+MCG \cite{melm_18}} & 36.9 & 22.9 & 8.4 & 32.9                               \\\hline
\multicolumn{2}{c|}{CAM+MCG \cite{CAM}} & 20.4 & 7.8 & 2.5 & 23.0                               \\\hline
\multicolumn{2}{c|}{SPN+MCG \cite{spn_iccv2017}} & 26.4 & 12.7 & 4.4 & 27.1                               \\\hline
\multicolumn{2}{c|}{PRM \cite{PRM}} & \sbest{44.3} & \sbest{26.8} & \sbest{9.0} & \sbest{37.6} \\\hline
\multicolumn{2}{c|}{Proposed } & \best{48.5} & \best{30.2} & \best{14.4} & \best{44.3} \\\hline
\end{tabular}
} \vspace{0.2em}
 \caption{Image-level supervised instance segmentation results on the PASCAL VOC 2012 val. set in terms of mean average precision (mAP\%) and Average Best Overlap(ABO). Our approach improves PRM \cite{PRM} with a relative gain of 17.8$\%$ in terms of ABO.}
\label{tab:ins_seg_pascal}
\vspace{-0.2cm}
\end{table}

\section{Conclusion}
We introduced a partially supervised setting for generic object counting in natural scenes, and proposed two novel frameworks to enable object counting under this challenging setting. Our  frameworks were built on a novel dual-branch architecture having an image classification branch, and a density branch that  estimates  the object count in an image. Our first framework (the LC framework) requires only  lower-count supervision, and hence reduces the  annotation cost due to large numbers of instances in an image. As an extension,  the second framework (the RLC framework) uses lower-count supervision only for a subset of object categories,  and hence further reduces the annotation cost due to large numbers of object categories appearing in natural scene datasets. To the best of our knowledge, we are the first  to propose image-level lower-count supervised training for a  density map, and  demonstrate the applicability of density maps in image-level supervised instance segmentation. Thorough experiments were performed on three challenging datasets (COCO, Visual Genome and PASCAL VOC) to evaluate the category-specific and category-independent object counting performance of the proposed method. The evaluations demonstrated that the proposed frameworks perform on par with approaches using  higher levels of supervision. 
\vspace{-0.5cm}
{
\bibliographystyle{ieee}
\bibliography{main}
}
\vspace{-2em}
\begin{IEEEbiography}[{\includegraphics[width=1in,height=1.25in,clip,keepaspectratio]{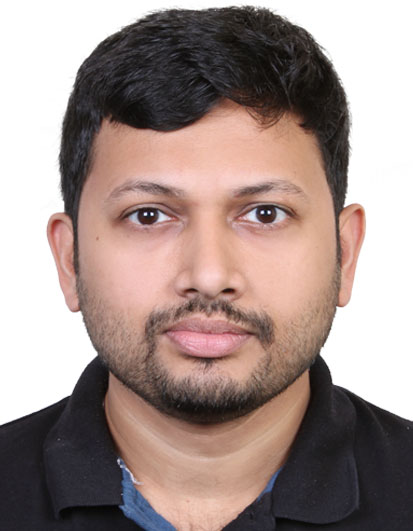}}]{Hisham Cholakkal}  received the Ph.D. degree from Nanyang Technological University, Singapore, and the M.Tech. degree from IIT Guwahati, India.  He is  an Assistant Professor  at MBZ University of Artificial Intelligence (MBZUAI), UAE.  Previously, from 2018 to 2020 he worked as a Research Scientist at Inception Institute of Artificial Intelligence, UAE and from 
2016 to 2018 he worked as a Senior Technical Lead at Mercedes-Benz R\&D India.   He has also worked as a researcher at BEL-Central Research Lab, India and Advanced Digital Sciences Center, Singapore. He has served as a program committee member for several top conferences, including CVPR, ICCV, NeurIPS and ECCV. His research interests include  object detection,  instance segmentation, object counting and weakly supervised learning.  
\end{IEEEbiography}
\vspace{-3em}
\begin{IEEEbiography}[{\includegraphics[width=1in,height=1.25in,clip,keepaspectratio]{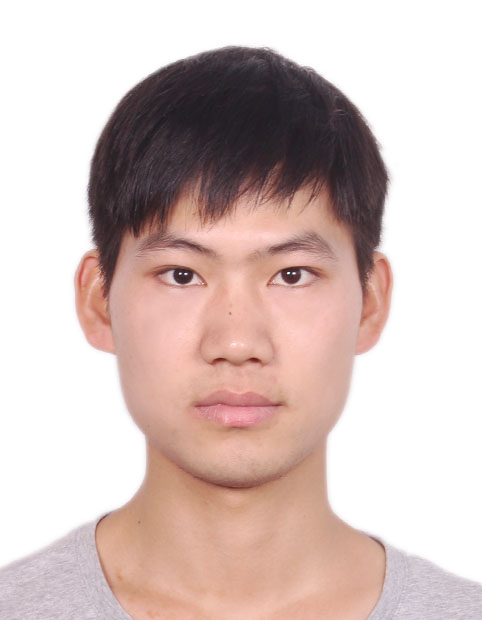}}] {Guolei Sun} is a PhD candidate at ETHZ, Switzerland, in Prof. Luc Van Gool’s Computer Vision Lab since 2019.  He received master degree in Computer Science from King Abdullah University of Science and Technology (KAUST) in 2018. From 2018 to 2019, he worked as research engineer at Inception Institute of Artificial Intelligence, UAE.
His research interests lie in deep learning for classification, semantic/instance segmentation, object counting and weakly supervised learning.
\end{IEEEbiography}
\vspace{-3em}
\vfill
\begin{IEEEbiography}[{\includegraphics[width=1in,height=1.25in,clip,keepaspectratio]{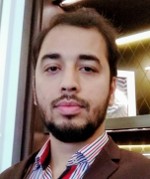}}]{Salman H. Khan}
received the Ph.D. degree from The University of Western Australia, in 2016. His Ph.D. thesis received an honorable mention on the Dean’s List Award. From 2016 to 2018, he was a Research Scientist with Data61, CSIRO. He then worked as a Senior Scientist at Inception Institute of Artificial Intelligence from 2018 to 2020.  He has been an Assistant Professor with MBZ University of AI, since 2020, and an Adjunct Faculty at Australian National University, since 2016. He has served as a program committee member for several premier conferences, including CVPR, NeurIPS, ICCV, and ECCV. His research interests include deep learning, computer vision and pattern recognition.
\end{IEEEbiography}
\vspace{-3em}
\begin{IEEEbiography}[{\includegraphics[width=1in,height=1.25in,clip,keepaspectratio]{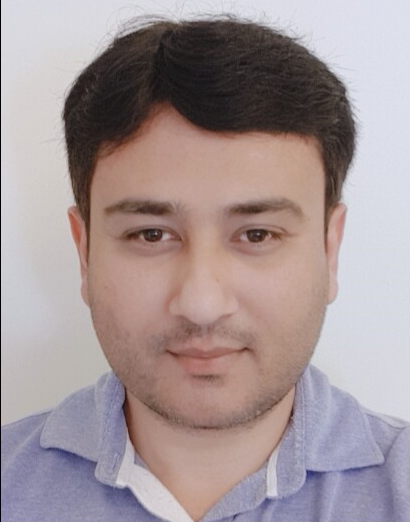}}]{Fahad Shahbaz Khan} is an Associate Professor at  MBZ University of AI, Abu Dhabi, United Arab Emirates and
Linkoping University, Sweden. From 2018 to 2020 he worked as 
 a Lead Scientist at the Inception Institute of Artificial Intelligence (IIAI),
Abu Dhabi, United Arab Emirates. He received the M.Sc. degree in Intelligent Systems Design from Chalmers University of Technology, Sweden and a Ph.D. degree in Computer Vision from Autonomous University of Barcelona, Spain. He has achieved top ranks on various international challenges (Visual Object Tracking VOT: 1st 2014 and 2018, 2nd 2015, 1st 2016; VOT-TIR: 1st 2015 and 2016; OpenCV Tracking: 1st 2015; 1st PASCAL VOC 2010). His
research interests include a wide range of topics within computer vision and machine learning, such as object recognition, object detection, action recognition and visual tracking.  He has published articles in high-impact computer vision journals and conferences in these areas.
\end{IEEEbiography}
\vspace{-3em}
\begin{IEEEbiography}[{\includegraphics[width=1in,height=1.25in,clip,keepaspectratio]{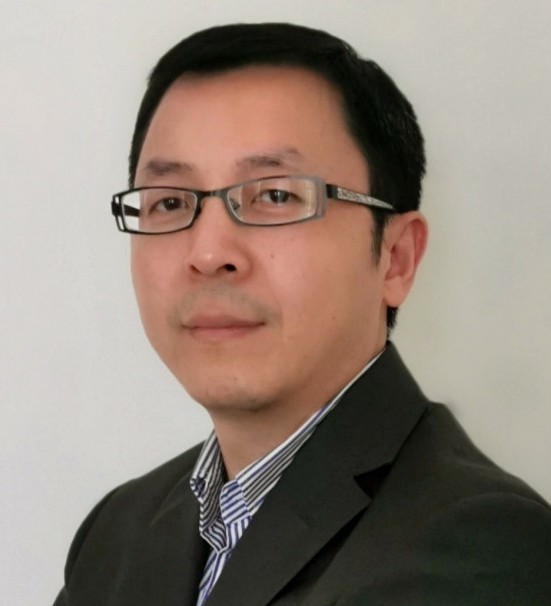}}]{Ling Shao}
is the CEO of the Inception Institute of Artificial Intelligence, and the Executive Vice President and Provost of the Mohamed bin Zayed University of Artificial Intelligence, Abu Dhabi, UAE. He was Chair Professor and Director of the Artificial Intelligence Laboratory at the University of East Anglia, Norwich, UK. He received the B.Eng. degree in Electronic and Information Engineering from the University of Science and Technology of China (USTC), the M.Sc degree in Medical Image Analysis and the PhD degree in Computer Vision at the Robotics Research Group from the University of Oxford. His research interests include Computer Vision, Deep Learning/Machine Learning, Multimedia, and Image/Video Processing. He has published over 300 papers at top venues such as TPAMI, TIP, IJCV, ICCV, CVPR, ECCV, etc.
\end{IEEEbiography}
\vspace{-3em}
\begin{IEEEbiography}[{\includegraphics[width=1in,height=1.25in,clip,keepaspectratio]{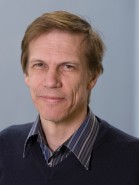}}] {Luc Van Gool} got a degree in electromechanical engineering at the Katholieke Universiteit Leu- ven in 1981. Currently, he is professor at the Katholieke Universiteit Leuven, Belgium, and the ETHZ, Switzerland. He leads computer vision research at both places. His main interests include 3D reconstruction and modeling, object recognition, and tracking, and currently especially their confluence in the creation of autonomous cars. On the latter subject, he leads a large-scale project funded by Toyota. He has authored over 300 papers in this field. He has been a program chair or general chair of several major computer vision conferences. He received several Best Paper awards, incl. a David Marr prize. In 2015, he received the 5-yearly excellence prize of the Flemish Fund for Scientific Research and, in 2016, a Koenderink Award. In 2017 he was nominated one of the main Tech Pioneers in Belgium by business journal ‘De Tijd’, and one of the 100 Digital Shapers of 2017 by Digitalswitzerland. He is a co-founder of 10 spin-off companies.
\end{IEEEbiography}
\vfill
\end{document}